\documentclass[11pt]{article}

  \usepackage[final]{acl}

\usepackage[T1]{fontenc}
\usepackage[utf8]{inputenc}
\usepackage{times}
\usepackage{latexsym}
\usepackage{microtype}
\IfFileExists{inconsolata.sty}{\usepackage{inconsolata}}{}
\usepackage{amsmath,amssymb,amsthm,mathtools}
\usepackage{booktabs}
\usepackage{array}
\usepackage{multirow}
\usepackage{graphicx}
\usepackage{xcolor}
\usepackage{enumitem}
\usepackage{url}
\usepackage{hyperref}
\usepackage{cleveref}

\hypersetup{
  colorlinks=true,
  linkcolor=blue,
  citecolor=blue,
  urlcolor=blue
}

\newtheorem{theorem}{Theorem}
\newtheorem{proposition}[theorem]{Proposition}
\newtheorem{corollary}[theorem]{Corollary}
\newtheorem{lemma}[theorem]{Lemma}
\theoremstyle{definition}
\newtheorem{definition}[theorem]{Definition}
\newtheorem{assumption}[theorem]{Assumption}
\theoremstyle{remark}

\newcommand{\E}{\mathbb{E}}
\newcommand{\Prob}{\mathbb{P}}
\newcommand{\Var}{\operatorname{Var}}
\newcommand{\KL}{\operatorname{KL}}
\newcommand{\TV}{\operatorname{TV}}
\newcommand{\chisq}{\chi^2}
\newcommand{\ind}{\mathbf{1}}
\newcommand{\Normal}{\mathcal{N}}
\newcommand{\calY}{\mathcal{Y}}

\newcommand{\dd}{\,\mathrm{d}}
\newcommand{\asympdist}{\rightsquigarrow}

\newcommand{\testlevel}{\tau}
\newcommand{\dcal}{\delta_{\mathrm{cal}}}
\newcommand{\dcert}{\delta_{\mathrm{cert}}}

\title{When Is Benchmark Contamination Detectable?\\
Information Limits and Power-Calibrated Audits}

\author{
\textbf{Ibne Farabi Shihab}\thanks{Equal contribution.}\thanks{Corresponding author: \texttt{ishihab@iastate.edu}.}\textsuperscript{1}
\and
\textbf{Sanjeda Akter}\footnotemark[1]\textsuperscript{1}
\and
\textbf{Anuj Sharma}\textsuperscript{2}
\\[2pt]
\textsuperscript{1}Department of Computer Science, Iowa State University \\
\textsuperscript{2}Department of Civil, Construction \& Environmental Engineering, Iowa State University \\
\texttt{ishihab@iastate.edu}
}
\date{}

\begin{document}
\maketitle

\begin{abstract}
Behavioral contamination detectors can return ``no evidence'' either because a
benchmark is clean or because the audit has little power. We formalize the
distinction for a benchmark in which an unknown fraction $\alpha$ of items was
seen during training: with matched clean and seen controls the behavioral
channel is the sparse mixture $Q_\alpha=(1-\alpha)P_0+\alpha P_1$, and an exact
second-moment argument shows detectability is governed by $\alpha\rho\sqrt m$,
where $\rho^2=\chisq(P_1\Vert P_0)$ measures behavioral separability. Any
scalar detector reduces to its efficacy
$e_f=|\E_1f-\E_0f|/\sqrt{\Var_0(f)}\le\rho$, which is estimable from controls
before the audit is run; a separate sample-split certificate lower-bounds
$\alpha$ distribution-free, without an orientation assumption. Our empirical
finding is two-sided. Frozen calibration efficacy predicts held-out power
\emph{curves} ($R^2=0.83$--$0.98$ over six exact-permutation channels), but the
efficacy-only Gaussian \emph{budget} is miscalibrated at the small $m$ it
prescribes, failing in $9/9$ gate-passing channels even though efficacy itself
transports---the inversion breaks, not the calibration. A predeclared two-stage
planner that simulates the deployed test repairs the budgets, is uniformly
conservative, and abstains where its probe does not transport. The certificate
is valid but vacuous at audit scale, and a five-seed paired injection study
recovers a mechanism ordering (verbatim $>$ paraphrase $>$ surface) in which
the apparent answer-only signal is baseline drift. We report the audit contract
and its failures together: a non-rejection is interpretable only alongside the
efficacy, budget, and validity gates that produced it.
\end{abstract}

\section{Introduction}

Public evaluation sets routinely appear in web-scale training corpora.  Exact
or semantic exposure can inflate a model's benchmark score without improving
the capability the benchmark was intended to measure
\citep{magar2022data,dong2024generalization}.  A growing literature therefore
tries to infer exposure from token likelihoods, perturbations, output order, or
elicited recall \citep{mattern2023membership,shi2024detecting,
oren2024proving,golchin2025dcq}.  Yet empirical studies find that detectors
often disagree, fail under distribution shift, or can be evaded
\citep{duan2024mia,fu2025does,meeus2025sok,samuel2025towards,
dekoninck2024evading,wang2026fragility}.  A 25-model study documents this
reliability gap---only 201 of 335 audit outcomes correct, with shift false
positives and underpowered post-hoc inference
\citep{zarzecki2026reliability}; it establishes the empirical failure, and our
question is whether a calibrated channel can predict it before an audit is
run---absent that, a non-rejection is hard to interpret.

The missing object is \emph{audit power}: how many independent items must be
probed to detect a small exposed fraction, how does the answer depend on the
model's observable memory of a seen item, and can a detector state, with
controlled error, which fractions it rules out?  Dataset-level methods
aggregate membership scores into significance tests
\citep{maini2024dataset,zhang2024pacost,puerto2025scaling}, ConStat estimates
performance inflation \citep{dekoninck2024constat}, and FTD controls a false
discovery rate \citep{zhang2026controllable}---important but different
deliverables: a valid $p$-value does not say which exposure fractions the
audit had power to detect, and inflation is not the fraction exposed.

We study the benchmark-level hypothesis $H_0:\alpha=0$ versus
$H_1:\alpha>0$,
where $\alpha$ is the fraction of benchmark items exposed during training.  A
probe outcome $Y$ can contain a scalar loss, a vector of token statistics, a
set of stochastic generations, or a black-box answer.  Conditional on an item
being clean or seen, its outcome follows $P_0$ or $P_1$.  Uniformly sampling
benchmark items gives the mixture
\begin{equation}
  Q_\alpha=(1-\alpha)P_0+\alpha P_1.
  \label{eq:mixture}
\end{equation}
This assumption is substantive---it requires matched clean controls and rules
out unmodeled spillover---and we state it visibly, test its implications, and
show what fails without it.

\paragraph{Contributions.}
\begin{enumerate}[leftmargin=*,itemsep=2pt,topsep=3pt]
  \item We connect behavioral contamination audits to classical sparse-mixture
  detection \citep{ingster1997problems,donoho2004higher,
  cai2011heterogeneous,cai2014optimal}, instantiating a sharp second-moment
  limit in the observable $\rho^2=\chisq(P_1\Vert P_0)$ for Bernoulli exposure
  and fixed contaminated subsets; we claim the audit interpretation, not a new
  mixture boundary.

  \item We operationalize that limit for arbitrary contamination scores:
  the efficacy $e_f\le\rho$ of the oriented mean-score audit built from
  $f$ yields a prospective \emph{local-asymptotic} power prediction and a
  finite-sample lower bound on channel separability.  Power \emph{curves}
  transport across splits ($R^2$ up to $0.99$); the efficacy-only Gaussian
  \emph{planner} is miscalibrated at the small budgets it prescribes, and
  we report that failure with its diagnosis rather than a pooled scaling
  law (\S\ref{sec:nontransport}).

  \item We provide a sample-split audit reporting a distribution-free lower
  confidence bound on the exposed fraction $\alpha$, with coverage for any
  independently learned bounded score, even one accidentally reversed.

  \item We design complementary evaluations: checkpoint-matched channels,
  gate-passing same-corpus continued-pretraining channels with direct
  budget trials, and causal injections with five paired clean-continuation
  seeds.  Primary falsifiable endpoints: prospective power prediction,
  efficacy transport (tested; efficacy itself largely transports, but the
  Gaussian efficacy-to-budget \emph{inversion} fails at prescribed
  budgets), and coverage.
\end{enumerate}

\section{Problem Formulation}
\label{sec:setup}

\subsection{From benchmark items to a behavioral channel}

Fix a trained language model, a benchmark population, an access regime, and a
predeclared probing procedure.  Let $Z\in\{0,1\}$ indicate whether a
uniformly sampled benchmark item was exposed during training.  The complete
outcome from that item is $Y\in\calY$.  We treat all repeated decodes,
perturbations, and token-level features from the \emph{same} item as one joint
outcome; counting them as independent probes would artificially inflate $m$.

\begin{assumption}[Matched partial-contamination channel]
\label{ass:channel}
For the audited model and probe, $Y\mid Z=0\sim P_0$ and
$Y\mid Z=1\sim P_1$.  Probe items are sampled independently and uniformly,
$Z\sim\operatorname{Bernoulli}(\alpha)$, and the conditional laws $P_0,P_1$
do not change with $\alpha$ over the range being audited.
\end{assumption}

The last clause is a local no-spillover/stability condition: it does not claim
training is literally itemwise, but specifies when a clean reference channel
transports to the suspect model.  Clean-control and held-out items measure
violations in \cref{sec:experiments}; heterogeneous and adaptive versions are
in \cref{app:extensions}.

Under \cref{ass:channel}, $m$ outcomes follow $P_0^{\otimes m}$ under $H_0$ and
$Q_\alpha^{\otimes m}$ under $H_1$.  A possibly randomized audit is a
measurable $\phi_m:\calY^m\to[0,1]$, where $\phi_m=1$ means ``detect
contamination,'' with type-I error $a_m(\phi_m)=\E_0[\phi_m]$, type-II error
$b_m(\phi_m,\alpha)=\E_\alpha[1-\phi_m]$, and power
$\pi_m(\phi_m,\alpha)=1-b_m(\phi_m,\alpha)$.
We call an audit level $\testlevel$ if $a_m\le\testlevel$.

\subsection{Why contamination fraction alone is insufficient}

There is no nontrivial guarantee that depends only on $(\alpha,m)$: setting
$P_1=P_0$ makes the clean and contaminated transcript laws identical
($a_m+b_m=1$ for every test), representing exposure that leaves no trace in
the chosen access channel, and an unconstrained clean reference law lets any
observed law be relabeled as clean.  A universal $\Theta(m^{-1/2})$ threshold
therefore needs an explicit informativeness parameter.

\begin{definition}[Behavioral separability]
\label{def:rho}
Assume $P_1\ll P_0$, let $r=\dd P_1/\dd P_0$, and define
\begin{align}
  \rho^2
  &:=\chisq(P_1\Vert P_0)\nonumber\\
  &=\E_0[(r(Y)-1)^2].
  \label{eq:rho}
\end{align}
\end{definition}

$\rho$ depends jointly on the model, mechanism, benchmark distribution, probe,
and access regime: zero when the probe carries no membership signal, large
when seen items behave in ways rare under $P_0$, and infinite off-support
(where a single witness beats the regular rate).  Our matching result concerns
the practically common overlapping, finite-$\rho$ regime.

\section{Fundamental Detection Limit}
\label{sec:theory}

\subsection{A finite-sample second-moment bound}

For compactness, define the second-moment radius
\begin{equation}
 C_m(\alpha,\rho)
 :=\frac12\left\{(1+\alpha^2\rho^2)^m-1\right\}^{1/2}.
\label{eq:second-moment-radius}
\end{equation}

\begin{theorem}[Power limit for partial contamination]
\label{thm:lower}
Under \cref{ass:channel}, suppose $P_1\ll P_0$ and
$\rho^2<\infty$.  For every $\alpha\in[0,1]$, every $m\ge1$, and every
possibly randomized test $\phi_m$,
\begin{align}
 a_m(\phi_m)+b_m(\phi_m,\alpha)
 &\ge 1-\min\{1,C_m(\alpha,\rho)\}.
 \label{eq:error-lower}
\end{align}
Consequently, every level-$\testlevel$ audit satisfies
\begin{align}
 \pi_m(\phi_m,\alpha)
 &\le \min\{1,\testlevel+\min\{1,C_m(\alpha,\rho)\}\}.
 \label{eq:power-upper}
\end{align}
\end{theorem}

The calculation is exact up to the final TV inequality
($\chisq(Q_\alpha^{\otimes m}\Vert P_0^{\otimes m})=(1+\alpha^2\rho^2)^m-1$,
then $\TV\le\frac12\sqrt{\chisq}$), sharper than passing through KL and
Pinsker; all steps are in \cref{app:proof-lower}.

The same limit covers the realistic design in which an auditor evaluates
every item of a benchmark with a \emph{fixed, unknown} contaminated subset of
size $k$: \cref{lem:fixed-subset} (stated and proved in
\cref{app:proof-fixed}) gives the analogous bound with weak-signal boundary
$k\rho/\sqrt m=\alpha\rho\sqrt m=\Theta(1)$ for $k=\alpha m$, the symmetric
oracle score attains it without knowing the subset, and the subset-averaged
bound is a Bayes and hence minimax lower bound.

\begin{corollary}[Regular detection boundary]
\label{cor:boundary}
For any sequence of channels and alternatives for which
$\alpha_m\rho_m\sqrt m\to0$, every asymptotic level-$\testlevel$ audit has
$\limsup_m\pi_m(\alpha_m)\le\testlevel$.  Thus it has no asymptotic power beyond
its false-positive rate.  A necessary order for nontrivial power is
\[
  \alpha_m=\Omega\!\left(\frac{1}{\rho_m\sqrt m}\right).
\]
\end{corollary}

The familiar $m^{-1/2}$ exponent appears only when $\rho_m$ remains bounded
away from zero and infinity; capacity alone does not upper-bound $\rho$
without a training-stability theorem, so the informal capacity step of prior
arguments is replaced by an observable channel quantity.

\subsection{A matching test}

The lower-bound scaling is attainable: the centered likelihood-ratio score
$g=r-1$ yields a test that is locally asymptotically optimal, attaining the
envelope $1-\Phi(z_{1-\testlevel}-h)$ at local alternatives
$\alpha_m=h/(\rho\sqrt m)$, with the oracle sample budget
$m_{\mathrm{oracle}}\approx((z_{1-\testlevel}+z_{1-\beta})/(\alpha\rho))^2$
(\cref{eq:oracle-sample-size}); every regular channel admits a test matching
the $\alpha\rho\sqrt m$ boundary (construction and LAN statement in
\cref{app:matching}).

\subsection{What an existing detector score can achieve}

Most practical methods output a scalar score rather than an estimated density
ratio.  The following proposition turns such a score into a power diagnostic.

\begin{proposition}[Score efficacy]
\label{prop:efficacy}
For any nonconstant $f\in L^2(P_0)$, let
\begin{align*}
 \Delta_f&=\E_1 f-\E_0f,&
 \sigma_f^2&=\Var_0(f),\\
 e_f&=\frac{|\Delta_f|}{\sigma_f}.&&
\end{align*}
Then $e_f\le\rho$, with equality if and only if
$f-\E_0f=c(r-1)$ $P_0$-almost surely for some nonzero constant $c$.
Moreover,
\[
 \E_{Q_\alpha}f-\E_0f=\alpha\Delta_f,
\]
so the standardized local shift of the mean-score audit is
$\alpha\sqrt m\,e_f$.
\end{proposition}

Thus AUROC and benchmark-level power answer different questions: two scores
with similar AUROC can have different low-$\alpha$ efficacy, weak
instance-level separation can still accumulate over a large benchmark, and
$e_f/\rho\in[0,1]$ measures how much local information a score uses.
For an oriented mean-score test, the operational planning equation is
\begin{equation}
  m_{\mathrm{target}}(f)
  \approx
  \left(
  \frac{z_{1-\testlevel}+z_{1-\beta}}
       {\alpha e_f}
  \right)^2.
  \label{eq:score-sample-size}
\end{equation}
Unlike \cref{eq:oracle-sample-size}, this quantity is estimable from matched
controls and refers to the chosen detector, not an unattained channel
oracle.  Two scope warnings, borne out empirically: $e_f$ summarizes the
local power of the \emph{oriented mean-score test} built from $f$, not of
every detector computable from $f$; and \cref{eq:score-sample-size} is a
first-order Gaussian inversion, inheriting every finite-sample defect of
the local approximation (discreteness, higher moments, alternative
variance, test conservatism)---\S\ref{sec:nontransport} measures exactly
this failure at small $m$.

A companion variance-adaptive finite-sample floor
(\cref{cor:efficacy-floor}, \cref{app:floor}) converts calibration samples of
a bounded score into a certified lower bound
$\rho\ge e_f\ge\underline e_f$ that holds with probability $1-\dcal$, using
the empirical-Bernstein standard-deviation radius of
\citet{maurer2009empirical}.  Substituting $\underline e_f$ into
\cref{eq:score-sample-size} is conservative about calibration uncertainty but
remains a local-asymptotic planning calculation, not a finite-sample power
guarantee.

\subsection{Adaptive and heterogeneous probes}

An auditor may choose the next prompt after observing previous answers.
Adaptivity can allocate queries toward more informative probes but cannot
exceed a cumulative KL information budget of
$\tfrac12\sum_t\log(1+\alpha^2\rho_t^2)$ over fresh items
(\cref{prop:adaptive,app:extensions}); independent heterogeneous probes admit
exact chi-square tensorization, with local information
$\alpha^2\sum_i\rho_i^2$.  The precise statement and the KL/Pinsker power
bound appear in \cref{app:extensions}.

\section{A Power-Calibrated Audit}
\label{sec:method}

\subsection{Learning an efficient score}

The oracle $r=p_1/p_0$ is unknown, so we learn a probabilistic classifier
$s_\theta(y)\approx\Prob(Z=1\mid Y=y)$ from labeled, matched clean and seen
controls under balanced sampling; the primary audit uses the clipped score
$f_\theta(y)\in[0,1]$ for stability and finite-sample certification, with the
unclipped implied ratio $s_\theta/(1-s_\theta)$ kept only as a cross-fitted
diagnostic, never as an oracle estimate of $\rho$.

To prevent optimistic power estimates, data are divided by original document
or benchmark item into a \emph{score-training split} (fits $f_\theta$), a
\emph{calibration split} (estimates $\E_0f$, $\E_1f$, $\sigma_f$, and
prospective budgets), and an untouched \emph{audit split} with fresh control
items for testing and confidence bounds.
Cross-fitting can rotate these roles, but every reported fold keeps the score
independent of its calibration and audit observations.  Existing scores
(loss, Min-K, ReCaLL, and others) enter exactly the same pipeline after their
orientation is chosen on the training split.

\subsection{Detection and prospective power}

Let $\widehat\Delta_f=\bar f_1-\bar f_0$ and
$\widehat\sigma_f$ be computed only on calibration controls.  The plug-in
efficacy
$\widehat e_f=|\widehat\Delta_f|/\widehat\sigma_f$ gives the point budget in
\cref{eq:score-sample-size}; replacing it by $\underline e_f$ gives a
one-sided, uncertainty-aware local budget.  The \emph{primary confirmatory
decision} is a one-sided two-sample permutation test of
$\widehat s_f(\bar f_A-\bar f_0^{\rm test})$, where
$\widehat s_f=\operatorname{sign}(\widehat\Delta_f)$ and all score choices and
orientations are frozen before the audit split is opened.  Conditional on
exchangeability of the pooled audit/clean scores under $H_0$ and on the frozen
score, the exact (randomized) permutation test controls type-I error at
$\testlevel=0.05$.  The deployed implementation is Monte-Carlo with $199$
permutations and the add-one estimator
$\hat p=(1+\#\{\text{perm}\ge\text{obs}\})/200$ (valid; conservative on a
$1/200$ grid); exchangeability is exactly what the validity gates
(\S\ref{sec:validity}) probe, so empirical size is reported per channel, never
asserted (the initial runs use a cheaper standardized-mean approximation,
making size a measured endpoint).  Our remaining empirical endpoints are power as a
function of $(\alpha,m)$ and its prediction from the frozen calibration
efficacy ($R^2$, MAE; \cref{tab:powerpred}); predicted versus observed local
sample-size targets for $80\%$ power (\cref{tab:efficacy}); and the smallest
detectable fraction at the available size (\cref{tab:sensitivity}).  Bootstrap
intervals resample original items, never individual tokens or decodes.

When the null mean is estimated from $n_0^{\rm test}$ fresh clean controls, the
local coordinate uses
$m_{\rm eff}=(m^{-1}+(n_0^{\rm test})^{-1})^{-1}$ in place of $m$, and every
main power curve accounts for it; equations below write $m$ for the externally
calibrated or $n_0^{\rm test}\gg m$ regime.

At $\testlevel=.05$ and $80\%$ power,
$z_{.95}+z_{.80}=2.486$.  Thus
\begin{equation}
  \alpha_{\min}(f,m)\approx\frac{2.486}{e_f\sqrt m}.
  \label{eq:detectable-alpha}
\end{equation}
\Cref{tab:sensitivity} makes the resulting sensitivity analysis usable when a
closed model does not permit matched control construction.

\subsection{A finite-sample lower confidence bound}

A test can say that contamination is present; a prevalence certificate says
how much, distribution-free for a bounded, independently learned score.

Theorem~\ref{thm:certificate} (stated formally in \cref{app:certificate},
proved in \cref{app:proof-certificate}) takes any $[0,1]$-valued score $f$
fixed independently of the data, three mutually independent samples---$m$
audit items, $n_0$ clean controls, $n_1$ seen controls---and returns
\begin{equation}
 \underline\alpha
 =\begin{cases}
 \min\bigl\{1,\max\bigl\{0,\tfrac{N_L}{D_U}\bigr\}\bigr\}, & D_U>0,\\[2pt]
 0, & D_U\le0,
 \end{cases}
 \label{eq:lcb}
\end{equation}
with $\varepsilon_{\rm cert}(n)=\sqrt{\log(6/\dcert)/(2n)}$,
where $N_L=(\bar f_A-\varepsilon_{\rm cert}(m))-(\bar f_0+\varepsilon_{\rm cert}(n_0))$
shrinks the observed audit--clean gap by \emph{both} sample radii and
$D_U=(\bar f_1+\varepsilon_{\rm cert}(n_1))-(\bar f_0-\varepsilon_{\rm cert}(n_0))$
inflates the control gap; then
$\Prob_\alpha\{\underline\alpha\le\alpha\}\ge1-\dcert$ by a simultaneous
three-sample Hoeffding argument.  The $D_U\le0$ guard matches the formal
statement in \cref{app:certificate}: without it, a reversed score can give
$N_L<0$ and $D_U<0$ with a spuriously positive ratio.  No orientation
assumption is needed (a reversed score yields $\underline\alpha=0$), the
bound is conservative by design, and rejecting when $\underline\alpha>0$ is a
fallback whose rejection rate we report beside the confirmatory permutation
test, not the source of the main power curves.

\subsection{Required validity checks}
\label{sec:validity}

The mathematics cannot rescue a mismatched control channel.  We therefore
make four diagnostics part of the audit contract: \textbf{blind separation}
(a text-only classifier must not separate the pools), \textbf{clean-channel
transport} (clean-control scores must not shift across runs),
\textbf{channel stability} (conditional score laws must be stable across
$\alpha$), and the \textbf{independence unit} (effective $m$ counts source-item
clusters, never near-duplicates or repeated decodes).  Full definitions and
thresholds are in \cref{app:gates}.

The interpretation is also predeclared: a pooled transport claim requires
median multiplicative sample-size error $\le2$ \emph{and} $\ge80\%$ of
held-out channels inside nominal $90\%$ prediction intervals; otherwise
the pooled scaling claim is withdrawn and calibration-to-audit
nontransport becomes the primary conclusion.  (The v5 intervals are
descriptive, so the second criterion is unevaluable as specified;
\S\ref{sec:nontransport}.)

\section{Experiments}
\label{sec:experiments}

\subsection{Research questions}

Four pre-specified questions: \textbf{RQ1 (prospective prediction)}---does
calibration-set efficacy predict held-out audit power and sample size?
\textbf{RQ2 (certification)}---does \cref{eq:lcb} attain nominal coverage,
and how conservative is it? \textbf{RQ3 (transport and
mechanism)}---how do mechanism, repetition, access, and post-training
alter $\widehat e_f$, its floor, and rankings? \textbf{RQ4 (scaling
diagnostic)}---does power collapse on
$\alpha\widehat e_f\sqrt{m_{\rm eff}}$ with slope near $-1/2$?

\subsection{Evaluation A: checkpoint-matched channels}

The design follows the checkpoint construction of
\citet{wang2026checkmia}: members from a window before checkpoint step $t$,
nonmembers from a matched unprocessed window, balanced on source, length,
packing, deduplication, and stream position, blind text-only baseline
first.  The completed runs use Pythia \citep{biderman2023pythia} and
GPT-Neo; in the six-channel audit the member and nonmember pools come from
\emph{different corpora}, so all six fail the blind gate
(\S\ref{sec:diagnostics})---the gate working as designed.  OLMo
\citep{groeneveld2024olmo} channels are planned, not run.

For each held-out channel, we create aggregate audit sets by sampling a
known fraction $\alpha$ of members and $1-\alpha$ nonmembers.  This tests
the statistical audit conditional on a real model channel; it does
\emph{not} claim the post-hoc mixture retrains the model or validates
no-spillover---that causal question is Evaluation B's.

\paragraph{Gate-passing channels (v3).}
A third audit randomly splits one corpus ($700{+}700$) and continues
pretraining each base model on the member half: pools text-exchangeable by
construction, membership causally induced, with \emph{direct} power trials
at the frozen budgets (all details in \cref{tab:gatesv3}).

\paragraph{Locked external validation.}
A locked validation on the open 25-model release of
\citet{zarzecki2026reliability}---efficacy calibrated without their audit
outcomes predicting their documented failures, budgets, and coverage, no
score selection or recalibration---is a planned extension, not a completed
result; the Zarzecki transport is not yet run.

\subsection{Evaluation B: controlled contamination injection}

We distinguish two contamination fractions that are easy to conflate: the
\emph{training injection fraction} $\beta_{\rm train}$ (items injected during
continuation over benchmark size), a causal intervention whose variation
requires independently trained models, and the \emph{audit exposed fraction}
$\alpha_{\rm audit}$ (exposed items over audit-set size), which can be varied
by post-hoc remixing conditional on one trained model.  Remixing
$\alpha_{\rm audit}$ measures audit power, not the causal effect of more
training contamination; they coincide only when the entire benchmark is
audited.

Starting from the clean EleutherAI/pythia-160m checkpoint (Pile-trained; see the
provenance caveat below), we continue training for a fixed token
and optimizer budget ($400$ AdamW steps, learning rate $5\times10^{-5}$). We
inject a $\beta_{\rm train}$ fraction of SQuAD benchmark items into a Pile
background corpus and audit against a disjoint held-out pool of the same
benchmark; the main injected condition uses $200$ injected and $200$ held-out
items. Total updates, token count, and order randomization are matched to a
clean ($\beta_{\rm train}=0$) run per training seed.

We cross the injection with four mechanisms at $k=4$ copies per injected item:
exact question--answer text; a \emph{validated paraphrase} (answer
preserved, overlap-capped, dual independent NLI judges; acceptance
$138/200=0.69$; pipeline in \cref{app:paraphrase}); a word-order shuffle
(a \emph{surface perturbation}); and answer/rationale-only exposure. Auditing both exposed and held-out items
makes spillover directly observable. The upgraded design runs \emph{five}
paired training seeds, each with a matched clean ($\beta_{\rm train}{=}0$)
continuation (same background, order, steps, optimizer, initialization); the
causal readout is the per-seed \emph{paired contrast} $D_s$ = (exposed$-$held
gap, contaminated) $-$ (same gap, clean) in clean-model units, with item-level
bootstrap intervals. Larger exposure counts remain future work.  ``Pile-trained'' does
not certify the base checkpoint never saw SQuAD, so we claim only the
\emph{incremental} exposure effect of the paired continuations---a
well-defined causal quantity even with prior exposure
(full caveat in \cref{app:provenance}).

\subsection{Probes and statistical protocol}

All reported experiments evaluate five gray-box probe families (mean loss,
zlib-normalized loss, Min-K\% Prob, Min-K\%++, neighborhood loss); every
``best probe'' is \emph{the probe with the highest calibration-split
efficacy}, fixed before audit power is computed.  ReCaLL, CAMIA-style,
black-box, and cross-fitted-combination probes are \emph{specified but not
run} (\cref{app:protocol}); no result depends on them.  The nominal confirmatory size is $\testlevel=0.05$, the
certificate miscoverage is $\dcert=0.05$, and the calibration-floor failure
probability is $\dcal=0.05$; these are distinct guarantees.  The target power
is $0.80$.  Probe counts are $m\in\{16,\dots,1024\}$ truncated at the number
of independent item clusters, with a fresh clean reference of
$n_0^{\rm test}=250$ items and $m_{\rm eff}$ reported per test.  Monte-Carlo
replication counts, interval conventions, onset-fitting rules, and the
Holm-adjusted secondary analyses are specified in \cref{app:protocol};
the transport decision uses the predeclared criteria of
\cref{sec:validity}.

\section{Results}
\label{sec:results}

\subsection{Calibration predicts held-out power}

\begin{figure*}[htbp]
    \centering
    \includegraphics[width=\linewidth]{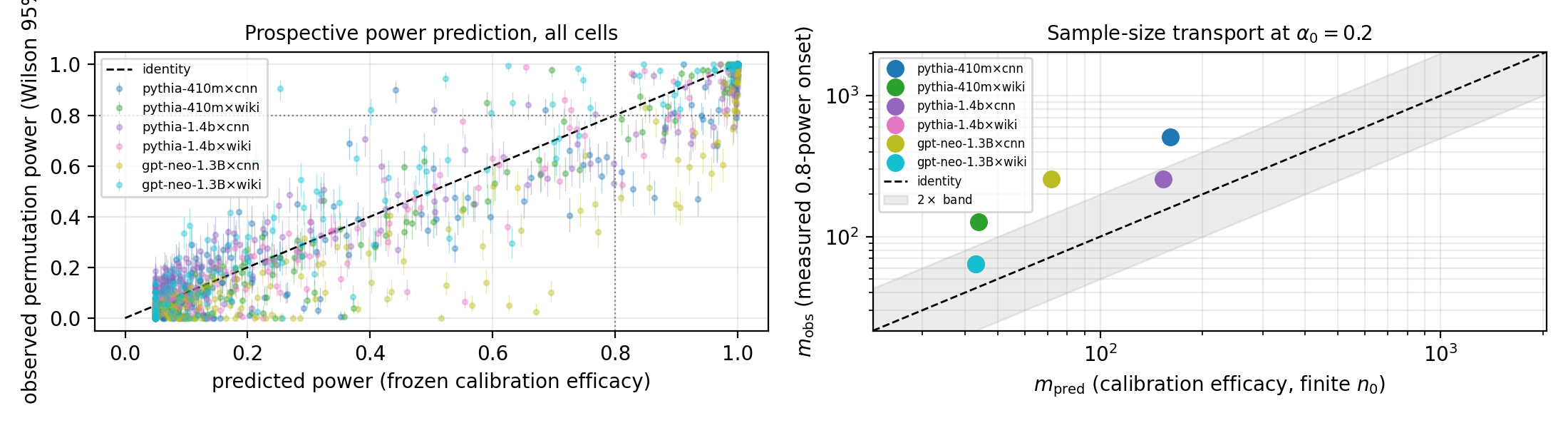}
    \caption{Upgraded six-channel audit, exact permutation test. Left:
    predicted vs.\ observed power, every (channel, probe, $\alpha$, $m$)
    cell, Wilson intervals. Right: $m_{\rm pred}$ vs.\ measured
    $m_{\rm obs}$ at $\alpha_0{=}0.2$ (log--log, $2\times$ band); the
    dyadic-grid onsets are \emph{interval-censored upper bounds}, so the
    apparent $1.5$--$3.6\times$ ratios are not resolved onset ratios (two
    of three cases are unresolved; \S\ref{sec:nontransport})---the
    censoring-free diagnosis comes from the v3 direct trials.}
    \label{fig:powerpred}
\end{figure*}

The initial audit freezes $\widehat e_f$ on the calibration split of one
$500{+}500$ Pythia-1.4b channel ($R^2=0.81$--$0.99$, MAE $0.02$--$0.10$;
\cref{tab:powerpred,tab:efficacy}; no fitted recalibration on the audit
split).  The primary evidence is an upgraded audit using the \emph{exact}
Monte-Carlo permutation test ($n_0{=}250$ reference, audits drawn without
replacement) over six model$\times$corpus channels (pythia-410m/1.4b,
GPT-Neo-1.3B; news and Wikipedia pools of $1{,}600{+}1{,}600$).
\Cref{fig:powerpred} shows every predicted-vs-observed cell with Wilson
intervals. Per channel (best probe): $R^2=0.83$--$0.98$, pooled MAE
$0.03$--$0.11$, slope $0.91$--$1.04$, intercept $-0.09$--$0.06$,
$0.8$-boundary accuracy $0.95$--$1.00$; in the informative $0.2$--$0.9$
band MAE grows to $0.06$--$0.25$---mid-band prediction is real but coarser.
Empirical size at $\alpha{=}0$ is $0.009$--$0.045$ (\cref{tab:gates}):
valid, somewhat conservative.  Scope, per our own protocol: these six
channels fail the blind gate (\S\ref{sec:diagnostics}), so their $R^2$
validates power-curve prediction \emph{within a content-confounded
channel}; the confirmatory reading rests on the gate-passing v3 channels,
and accurate curves do \emph{not} license plug-in budgets
(\S\ref{sec:nontransport}); values in the
\cref{fig:powerpred} caption.

\subsection{Plug-in budgets fail: the efficacy-only planner is miscalibrated at small \texorpdfstring{$m$}{m}}
\label{sec:nontransport}

Accurate power \emph{curves} do not imply usable plug-in \emph{budgets}.
The pre-specified criterion (\S\ref{sec:validity}) fails on the initial
channel: \cref{tab:efficacy} reports \emph{efficacy-transport ratios}
$\exp(2|\log\widehat e_f^{\rm cal}-\log\widehat e_f^{\rm test}|)$ with
median $2.23\times$ and $3/5$ channels above the $2\times$ tolerance, so
\textbf{we withdraw the pooled calibration-to-audit scaling claim}
(tolerance fixed before the numbers were seen).  But with $800$-item
calibration halves the ratios shrink to $1.03$--$1.50$, and in the v3
gate-passing channels the \emph{raw} efficacy ratios are $1.02$--$1.35$
($1.04$--$1.81$ squared to the budget scale
$m\propto e_f^{-2}$)---within the $2\times$ tolerance.  Efficacy
itself largely transports; that is \emph{not} what breaks the budgets.

The decisive diagnosis comes from the v3 censoring-free direct trials
(\cref{tab:gatesv3}; the earlier dyadic-grid ``$0/3$'' claim is withdrawn
as interval-censored, arithmetic in \cref{tab:gates}).  At the frozen
$m_{\rm pred}$ the planner predicts $0.80$ by construction; observed power
is $0.67$, $0.56$, $0.62$, Wilson intervals excluding $0.80$ in all three
channels; re-running the \emph{same} plug-in with the audit-split efficacy
still predicts $0.58$, $0.77$, $0.68$---errors of both signs up to $0.21$
that dwarf the efficacy shift (on pythia-410m$\times$wiki the efficacies
nearly coincide, $4.62$/$4.31$, yet the plug-in predicts $0.77$ against
$0.56$).  The failure is \textbf{not} calibration-to-audit transport; it is
\emph{consistent with} the local Gaussian approximation breaking at the
small budgets it prescribes ($m=8$--$42$) plus a conservative exact test
(size $0.000$--$0.013$), but the causal decomposition is not isolated (no
skewness/tie or Gaussian-vs-Welch-vs-permutation ablation was run; open).  At
$m_{\rm infl}=1.40\times m_{\rm pred}$ (frozen before scoring; uncertainty
not propagated, disclosed): two clear failures, one unresolved ($0.76$
$[0.70,0.81]$).

The constructive repair is \emph{executed at scale}.  A
\textbf{finite-sample planner} resamples full $P_0/P_1$ score laws from
calibration data only and runs the deployed permutation test at each
candidate $m$.  The v4 pilot (\cref{tab:plannerv4}) beat the Gaussian
budget head-to-head but chose $m$ from pointwise bootstrap bounds and
retained only two channels.  The v5
protocol repairs both defects (\cref{tab:plannerv5}): $12$ same-corpus
channels, an equivalence blind gate ($9/12$ pass the predeclared one-sided
form, AUC upper CI $\le0.55$; $8/12$ pass the stricter two-sided
sensitivity gate, which additionally excludes exactly the channel where
the planner abstains; per-channel values in \cref{tab:v5gates}), a
two-stage planner (stage-A bootstrap candidate $m_A$ on one calibration
half; stage-B confirmation on the untouched half over the pre-specified
ladder $\{m_A,\lceil1.5m_A\rceil,2m_A\}$ with Bonferroni-corrected
one-sided bounds), and the \textbf{prespecified transport endpoint} on the
audit half (estimated onset $\widehat m_\star$, multiplicative errors,
$90\%$ bootstrap quantile intervals).
\emph{Validity scope, stated exactly:} the Bonferroni step makes stage B
valid against \emph{rung selection} conditional on the empirical cal-B
distribution ($175$ clean clusters resampled with replacement to
$n_0{=}250$), excluding score-law uncertainty, cluster reuse, and
cal-to-audit transport (measured by the audit endpoint, not certified);
$\widehat m_\star$ is a Monte-Carlo onset estimate ($200$ trials), the
$90\%$ intervals are descriptive ($B{=}30$, probe selection not re-run),
the ratios and $8/8$ coverage inherit that noise
($\widehat\pi(m_{\rm plan})=0.785$ at $m_{\rm plan}=\widehat m_\star=56$
on pythia-70m$\times$cnn), and the prespecified $\ge80\%$
interval-coverage criterion is \emph{unevaluable as specified} and not
claimed.
Results: the Gaussian budget \emph{clearly fails in $9/9$} channels
(power $0.01$--$0.74$ at $m_{\rm pred}$, every Wilson interval excluding
$0.80$; $m_{\rm pred}/\widehat m_\star=0.54$--$0.86$, median $0.59$); the
planner is uniformly conservative
($m_{\rm plan}/\widehat m_\star=1.00$--$2.00$, median $1.36$), delivering
$0.785$--$0.92$ power ($\ge0.80$ in $6/8$, two unresolved, zero clear
failures; $90\%$ intervals cover $\widehat m_\star$ in $8/8$).
On pythia-70m$\times$wiki the planner \emph{abstains} (stage A never
certifies $0.80$) while the Gaussian budget prescribes $m{=}22$ at $0.01$
power---abstention is the designed behavior, and the two-sided gate
excludes this channel outright.  The certificate (below) makes no
planning assumption and the injection ordering (\S\ref{sec:diagnostics})
stands.

\subsection{Certificate coverage: valid, and vacuous at audit scale}
\label{sec:certcov}

The scaled-slack heuristic's $99.0$--$99.5\%$ endpoints
(\cref{tab:coverage}) are not validation of the distribution-free
certificate, which never fires at $m{=}512$ (zero false, zero nonzero in
all six channels; coverage vacuously $100\%$): certified
$\alpha{=}0.1$--$0.2$ needs balanced groups of
$\sim3{,}500$--$79{,}000$ items at measured margins.  A nonzero
$\underline\alpha$ is trustworthy; a zero is uninformative---at audit
sizes the certificate is a null instrument, reported as a limitation
(full analysis: \cref{app:certcov-full,app:nonvacuous,app:scaling-diag}).

\subsection{Mechanism, access, and validity diagnostics}
\label{sec:diagnostics}

The initial three-seed injections already show the mechanism decay
($0.741$/$0.343$/$0.150$ exact/surface/answer-only, $3/3$ seeds;
\cref{tab:coverage}).

The five-seed \emph{paired-contrast} rerun strengthens this, using
\emph{one common frozen probe} (mean NLL, the modal calibration winner)
since the per-cell best probe varies by seed and mechanism.  The contrasts $D_s$ (item-bootstrap
$95\%$ intervals excluding zero in all $15$ seed$\times$mechanism cells)
are $1.01\pm0.07$ exact, $0.70\pm0.09$ paraphrase, $0.44\pm0.04$ surface,
ordering intact in $5/5$ seeds; answer-only has $D_s=-0.01\pm0.07$ with
every per-seed interval covering zero---its apparent raw efficacy
($0.19$) is matched by the clean continuation (baseline drift, the
artifact the paired design exists to catch), and it is where
exchangeability degrades (pooled size $0.087$ vs.\ $0.013$--$0.058$).
We claim only ``indistinguishable from clean continuation at $n{=}200$,
$k{=}4$'' for answer-only; the exact certificate at injection scale never
fires (zero false, zero nonzero, coverage $1.0$), matching
\cref{app:nonvacuous}.

Every per-seed $D_s$, interval, size-gate value, and adjacent paired
difference is in \cref{tab:perseed}; on the common probe all three
adjacent differences ($0.31\pm0.12$, $0.26\pm0.10$, $0.45\pm0.08$) are
positive in $5/5$ seeds (sign-test $p=0.031$ each; Holm
$p_{\rm adj}=0.094$: a consistent descriptive pattern, not individually
significant after correction, and---as the arms are not
dose-matched---not a causal mechanism comparison).  The exposure-effect
claim is scoped to pythia-160m, SQuAD, $k{=}4$, this protocol.

The exact-injection channel also compares plug-in predictions against
observed $0.8$-power onsets: per seed $m_{\rm pred}$ of $371$, $208$,
$301$ vs.\ $m_{\rm obs}$ of $512$, $256$, $512$ (dyadic grid, quantized
upward), within the
$2\times$ tolerance in all three seeds; surface and answer-only never
reach $0.8$ power on the tested grid---censored, not extrapolated.
Efficacies are estimates or lower bounds on $\rho$; the small-sample
floor vacuity at $n{=}200$ is in \cref{tab:floor-comparison}.

The validity gates are measured per channel (\cref{tab:gates}): every
checkpoint-free Evaluation~A channel \emph{fails} blind separation
(text-only AUC $0.95$---content-confounded separability, exactly what the
gate exposes), while the v3 same-corpus channels repair it (blind AUC
$0.48$--$0.54$; $4/6$ pass all gates, the two Pythia$\times$SQuAD channels
failing the size gate, \cref{tab:gatesv3}).  The v3 rule is not an
equivalence test; requiring upper CI $\le0.55$ also excludes
pythia-410m$\times$wiki (upper $0.579$), leaving the small-$m$ conclusion
on the two surviving channels ($0.67$, $0.62$, both excluding $0.80$).
Pure-memorization claims rest on the gate-passing v3 channels and the
Evaluation~B injections.

% Sparse-mixture detection, the $\alpha\rho\sqrt m$ regime, and
% mixture-proportion estimation are classical
% \citep{ingster1997problemsproblemsproblemsproblems,donoho2004higher,cai2011heterogeneous,cai2014optimal,lecam1986asymptoticasymptotic,vanderVaart1998asymptoticasymptotic,blanchard2010novelty,scott2015rate,ramaswamy2016mixture,rogan1978estimating};
% we claim audit deliverables, not new mixture theory.  The closest LLM
% audits appear with their estimands in \S1; shift fragility motivates the
% gates
% \citep{fu2025does,das2025blind,wang2026checkmia,puerto2025scaling,duan2024mia,meeus2025sok,wang2026fragility}.
\section{Related Work}
\label{sec:related}

Behavioral detectors infer exposure from likelihoods, perturbations,
output order, or elicited recall \citep[\emph{inter
alia}]{mattern2023membership,shi2024detecting,oren2024proving}, and
evaluations of them report a reliability gap rather than a ranking
\citep{duan2024mia,das2025blind,zarzecki2026reliability}: they
establish \emph{that} audits fail, not a quantity---measurable before
an audit---saying when they will. Dataset-level methods aggregate item
scores into $p$-values, inflation estimates, or FDR-controlled
decisions
\citep{maini2024dataset,dekoninck2024constat,zhang2026controllable};
none states which $\alpha$ the audit had power to detect. The
$\alpha\rho\sqrt m$ boundary and prevalence estimation are classical
\citep{ingster1997problems,cai2014optimal,scott2015rate}; we claim audit
deliverables, not new mixture theory. Appendix~\ref{app:related} gives
the full treatment, including a per-method comparison of estimands
(Table~\ref{tab:estimands}).

% Curves transport; Gaussian budgets fail; the planner materially improves
% the audited budgets.
\section{Discussion and Conclusion}

Frozen calibration efficacy predicts held-out audit power
($R^2=0.83$--$0.98$), but its Gaussian inversion into a budget does
not: the plug-in fails in $9/9$ gate-passing channels
($m_{\mathrm{pred}}/\widehat m_\star=0.54$--$0.86$) even though
efficacy itself transports (ratios $1.02$--$1.35$), so the culprit is
the local approximation at the small $m$ it prescribes, not
calibration drift. Simulating the deployed test instead yields
budgets that are uniformly conservative, deliver $0.785$--$0.92$
power, and abstain where the probe fails to transport. A
non-rejection is thus interpretable only with its efficacy, budget,
and gate values attached---and the prevalence certificate, though
valid, is vacuous at audit scale.

\section*{Limitations}

The mixture channel is an auditable modeling assumption, not a universal law
of training.  Contaminated examples can affect unexposed examples, and
post-training can alter the clean and seen channels.  Our stability diagnostics
can reveal large violations but cannot prove that a clean reference is
perfectly transportable.  The lower bound is pointwise in a chosen behavioral
access channel; it does not apply to direct corpus search, cryptographic
provenance, or support-separated watermark evidence.  The matching optimality
result is local and assumes finite chi-square divergence and a mild moment
condition.  The distribution-free prevalence certificate requires independent
matched controls and a score learned on separate data; using the audit set to
tune the score invalidates its coverage.  Such controls are generally
unavailable to an external auditor of a closed model, so neither efficacy nor
the exposed fraction is then identified without extra assumptions.  Finally,
contamination encompasses more than verbatim membership.  Semantic task
exposure, answer-only exposure, and distillation may require different labels
and channels, so results should not be transported between mechanisms without
validation.  Six specific gaps remain open and unexecuted: the v5 planner's stage-B
guarantee is conditional on the empirical cal-B distribution (a
population-level selection-valid planner would need to propagate
score-law uncertainty and cluster reuse, e.g.\ by nesting the bootstrap
over probe selection with a larger replicate budget); the prespecified
interval-coverage transport criterion is unevaluable with descriptive
intervals and is not claimed; the as-run size gate has the detection
rather than equivalence direction (the equivalence form leaves $4/12$
channels, \cref{tab:v5gates}); the cause of Gaussian miscalibration is
consistent with small-$m$ non-Gaussianity plus test conservatism but not
isolated by ablation; the mechanism ordering is not dose-matched (a
causal comparison requires retraining all arms on the $138$-item
paraphrase intersection with copy- and token-matched exposure); and no
locked external or modern-model validation (OLMo-class models, black-box
access, the Zarzecki study, or matched-counterpart designs such as
LLM Dataset Inference or PaCoST) has been run.  The certificate remains a
theoretical contribution: mathematically valid but firing on no reported
condition.  Where these gaps would change a claim, the claim is scoped
accordingly.

\section*{Ethical Considerations}

Contamination audits can improve scientific accountability, but membership
signals can also expose whether copyrighted, private, or sensitive text was
used for training.  Experiments should use licensed or public data, avoid
releasing reconstructable private examples, and report aggregate statistics.
The proposed lower bound must not be used to imply that undetected
contamination is acceptable.  Controlled contamination runs should be confined
to research models and should not publish contaminated checkpoints as
general-purpose models without prominent documentation.  All experimental
results, compute use, data licenses, and automated writing or coding assistance
must be disclosed in the venue's Responsible NLP checklist.

\bibliography{references}

\appendix

\section{Proofs}
\label{app:proofs}

\subsection{Proof of the finite-sample power limit}
\label{app:proof-lower}

Put $g=r-1$.  The one-observation likelihood ratio is
$L_\alpha=1+\alpha g$.  Because $\E_0g=0$ and $\E_0g^2=\rho^2$,
\[
 \E_0L_\alpha^2=1+\alpha^2\rho^2.
\]
The product likelihood ratio is
$L_\alpha^{(m)}=\prod_{i=1}^mL_\alpha(Y_i)$.  Independence therefore gives
\begin{align}
 1+\chisq(Q_\alpha^{\otimes m}\Vert P_0^{\otimes m})
 &=\E_0\{L_\alpha^{(m)}\}^2\nonumber\\
 &=\prod_{i=1}^m\E_0L_\alpha(Y_i)^2\nonumber\\
 &=(1+\alpha^2\rho^2)^m.
 \label{eq:product-chi-proof}
\end{align}

For two simple hypotheses, every randomized test satisfies
\[
 a_m+b_m
 \ge 1-\TV(P_0^{\otimes m},Q_\alpha^{\otimes m}).
\]
Indeed,
\begin{align*}
 a_m+b_m
 &=1-\{\E_{Q_\alpha^{\otimes m}}\phi_m
       -\E_{P_0^{\otimes m}}\phi_m\}\\
 &\ge1-\TV(P_0^{\otimes m},Q_\alpha^{\otimes m}),
\end{align*}
because $0\le\phi_m\le1$.  Cauchy--Schwarz applied to
$\TV(P,Q)=\tfrac12\E_Q|\dd P/\dd Q-1|$ gives
\[
 \TV(P,Q)\le\frac12\sqrt{\chisq(P\Vert Q)}.
\]
Combining this inequality with \cref{eq:product-chi-proof} and the trivial
bound $\TV\le1$ proves \cref{eq:error-lower}.

Finally,
\begin{align*}
 \pi_m(\alpha)-a_m
 &=\E_{Q_\alpha^{\otimes m}}\phi_m
   -\E_{P_0^{\otimes m}}\phi_m\\
 &\le\TV(P_0^{\otimes m},Q_\alpha^{\otimes m}).
\end{align*}
If $a_m\le\testlevel$, substitution of the same total-variation bound gives
\cref{eq:power-upper}. \qed

\subsection{Statement and proof for a fixed contaminated subset}
\label{app:proof-fixed}

\begin{lemma}[A fixed unknown contaminated subset]
\label{lem:fixed-subset}
Suppose exactly $k$ of the $m$ audited items are contaminated.  For
$S\subseteq[m]$, $|S|=k$, let
$P_S=\bigotimes_{i\in S}P_1\otimes\bigotimes_{i\notin S}P_0$, and let
$\bar b_{m,k}$ be the type-II error averaged uniformly over such subsets.
Then every test satisfies
\begin{align}
 a_m+\bar b_{m,k}
 &\ge 1-\min\{1,C_{m,k}^{\rm fix}\}, \label{eq:fixed-exact}\\
 C_{m,k}^{\rm fix}
 &:=\frac12\sqrt{\E(1+\rho^2)^J-1}, \nonumber\\
 \E(1+\rho^2)^J
 &\le\left(1+\frac{k\rho^2}{m}\right)^k. \label{eq:fixed-mgf}
\end{align}
where $J=|S\cap S'|$ for independent uniform $k$-subsets $S,S'$.
\end{lemma}

For achievability in a fixed regular channel, the local regime additionally
requires $k/m\to0$, $k\rho/\sqrt m\to h$, and
$\Var_1(r-1)<\infty$; there the symmetric oracle score
$\sum_i(r(Y_i)-1)$ attains the boundary without knowing the subset.

Let $\mathcal S_k=\{S\subseteq[m]:|S|=k\}$ and
$\overline P_{m,k}=|\mathcal S_k|^{-1}\sum_{S\in\mathcal S_k}P_S$.
Because type-II error is linear in the alternative,
\[
 \bar b_{m,k}=\E_{\overline P_{m,k}}(1-\phi_m).
\]
Under $P_0^{\otimes m}$, the likelihood ratio of $P_S$ is
$L_S=\prod_{i\in S}r(Y_i)$, so that of $\overline P_{m,k}$ is
$\overline L=\E_S L_S$.  Tonelli's theorem and independence give
\begin{align*}
 1+\chisq(\overline P_{m,k}\Vert P_0^{\otimes m})
 &=\E_0\overline L^2\\
 &=\E_{S,S'}\E_0[L_SL_{S'}]\\
 &=\E_{S,S'}(1+\rho^2)^{|S\cap S'|}.
\end{align*}
Indeed, an index in $S\cap S'$ contributes
$\E_0r^2=1+\rho^2$; every other selected index contributes
$\E_0r=1$.  Applying the testing-error/TV argument and the chi-square TV bound
from \cref{app:proof-lower} proves \cref{eq:fixed-exact}.

It remains to prove \cref{eq:fixed-mgf}.  Conditional on $S$, let
$a_i=1+\rho^2$ for $i\in S$ and $a_i=1$ otherwise.  Uniformly sampling $S'$
without replacement gives
\begin{align*}
 \E[(1+\rho^2)^J\mid S]
 &=\binom{m}{k}^{-1}
   \sum_{\substack{T\subseteq[m]\\|T|=k}}\prod_{i\in T}a_i.
\end{align*}
Maclaurin's inequality bounds this normalized elementary symmetric mean by
the $k$th power of the arithmetic mean:
\begin{align*}
 \E[(1+\rho^2)^J\mid S]
 &\le\left(\frac1m\sum_{i=1}^ma_i\right)^k\\
 &=\left(1+\frac{k\rho^2}{m}\right)^k.
\end{align*}
The bound is independent of $S$, proving \cref{eq:fixed-mgf}.  In particular,
if $k^2\rho^2/m\to0$, its right-hand side is at most
$\exp(k^2\rho^2/m)\to1$, so average testing power cannot exceed size
asymptotically.  Since $\sup_S b_m(S)\ge\bar b_{m,k}$, the same display is a
minimax lower bound over fixed subsets.

For achievability in a fixed channel, let
$Z_m=(\rho\sqrt m)^{-1}\sum_i g(Y_i)$ and suppose
$k/m\to0$, $k\rho/\sqrt m\to h<\infty$, and
$\Var_1(g)<\infty$.  Since
\[
 \E_1g=\E_0[r(r-1)]=\E_0(r^2-r)=\rho^2,
\]
$\E_{P_S}Z_m=k\rho/\sqrt m\to h$.  The centered sum over the $m-k$ clean
indices, divided by $\rho\sqrt m$, converges to $\Normal(0,1)$ by the ordinary
CLT because $k/m\to0$.  The centered sum over the $k$ member indices is
$o_{P_S}(1)$ after the same normalization, since its variance is
$k\Var_1(g)/(m\rho^2)\to0$.  Slutsky's theorem therefore gives
$Z_m\asympdist\Normal(h,1)$ under every $P_S$, while its null limit is
$\Normal(0,1)$.  The one-sided score test consequently attains power
$1-\Phi(z_{1-\testlevel}-h)$ without knowing $S$. \qed

\subsection{Proof of the detection-boundary corollary}

If $\alpha_m\rho_m\sqrt m\to0$, then
$m\alpha_m^2\rho_m^2\to0$.  Since $1+x\le e^x$ for $x\ge0$,
\[
 C_m(\alpha_m,\rho_m)
 \le\frac12\sqrt{\exp(m\alpha_m^2\rho_m^2)-1}
 \longrightarrow0.
\]
Apply \cref{eq:power-upper} and take the upper limit. \qed

\subsection{Local asymptotic optimality}
\label{app:proof-upper}

\begin{theorem}[Local asymptotic optimality; full statement]
\label{thm:upper}
Suppose $0<\rho^2<\infty$ and, for some $\epsilon>0$,
$\E_0|g|^{2+\epsilon}<\infty$.  Fix $h\ge0$ and let
$\alpha_m=h/(\rho\sqrt m)$ for all sufficiently large $m$.  Then
\begin{align}
 \log\frac{\dd Q_{\alpha_m}^{\otimes m}}
               {\dd P_0^{\otimes m}}
 &=hZ_m-\frac{h^2}{2}+o_{P_0}(1), \label{eq:lan}\\
 Z_m&\asympdist\Normal(0,1) \quad\text{under }H_0,\\
 Z_m&\asympdist\Normal(h,1) \quad\text{under }H_{1,m}.
\end{align}
The score test $\ind\{Z_m>z_{1-\testlevel}\}$ has asymptotic size
$\testlevel$ and power $1-\Phi(z_{1-\testlevel}-h)$.  Every test sequence
$\{\psi_m\}$ with $\limsup_m\E_0\psi_m\le\testlevel$ satisfies
\[
 \limsup_m\E_{\alpha_m}\psi_m
 \le 1-\Phi(z_{1-\testlevel}-h).
\]
\end{theorem}

\paragraph{Proof.}

Let $g=r-1$.  Because $r\ge0$, $g\ge-1$.  Also,
\[
 \E_0g=\int(p_1-p_0)\dd y=0,\qquad
 \E_0g^2=\rho^2.
\]
The central limit theorem gives
\[
 Z_m=\frac{1}{\rho\sqrt m}\sum_{i=1}^m g(Y_i)
 \asympdist\Normal(0,1)
\]
under $P_0^{\otimes m}$.

If $h=0$, the likelihood ratio is identically one and all stated limits are
immediate.  Hence assume $h>0$.
For $\alpha_m=h/(\rho\sqrt m)$, the log-likelihood ratio is
\[
 \Lambda_m
 =\sum_{i=1}^m\log(1+\alpha_m g(Y_i)).
\]
Let $\epsilon'=\min\{\epsilon,1\}$.  The moment condition and a union bound
imply, for every fixed $c>0$,
\begin{align*}
 &\Prob_0\!\left(\max_{i\le m}|\alpha_mg(Y_i)|>c\right)\\
 &\quad\le
 m\Prob_0\!\left(|g|>\frac{c}{\alpha_m}\right)\\
 &\le
 \frac{m\alpha_m^{2+\epsilon'}}
      {c^{2+\epsilon'}}\E_0|g|^{2+\epsilon'}
 \longrightarrow0.
\end{align*}
On the event $\max_i|\alpha_mg_i|\le1/2$, Taylor's theorem gives a constant
$C_{\epsilon'}$ such that
\begin{align*}
 \left|\log(1+x)-x+\frac{x^2}{2}\right|
 &\le C_{\epsilon'}|x|^{2+\epsilon'},\\[-2pt]
 &\hspace{-40pt}|x|\le1/2.
\end{align*}
Therefore the summed remainder is bounded in probability by
\begin{align*}
 &C_{\epsilon'}\alpha_m^{2+\epsilon'}
 \sum_{i=1}^m|g(Y_i)|^{2+\epsilon'}\\
 &\quad=O_{P_0}(m\alpha_m^{2+\epsilon'})
 =o_{P_0}(1).
\end{align*}
The weak law of large numbers also gives
$m^{-1}\sum_i g(Y_i)^2\to\rho^2$ in $P_0$ probability.  Combining these facts,
\begin{align*}
 \Lambda_m
 &=\alpha_m\sum_{i=1}^mg(Y_i)\\
 &\quad-\frac{\alpha_m^2}{2}\sum_{i=1}^mg(Y_i)^2
   +o_{P_0}(1)\\
 &=hZ_m-\frac{h^2}{2}+o_{P_0}(1),
\end{align*}
which is \cref{eq:lan}.

Under $P_0^{\otimes m}$, $(Z_m,\Lambda_m)$ converges jointly to
$(Z,hZ-h^2/2)$ for $Z\sim\Normal(0,1)$.  Since
$\E[\exp(hZ-h^2/2)]=1$, Le Cam's third lemma applies and gives
$Z_m\asympdist\Normal(h,1)$ under $Q_{\alpha_m}^{\otimes m}$.  Hence
\[
 \Prob_0(Z_m>z_{1-\testlevel})\to\testlevel
\]
and
\[
 \Prob_{\alpha_m}(Z_m>z_{1-\testlevel})
 \to1-\Phi(z_{1-\testlevel}-h).
\]

It remains to justify the power envelope.  For each fixed $m$, the
Neyman--Pearson lemma makes a possibly randomized threshold test of
$\Lambda_m$ most powerful for the simple alternative
$Q_{\alpha_m}^{\otimes m}$.  Let $\{\psi_m\}$ be any test sequence with
$\limsup_m\E_0\psi_m\le\testlevel$.  For every
$\delta\in(0,1-\testlevel)$, it eventually has size at
most $s=\testlevel+\delta$, so Neyman--Pearson bounds its power by that of the
likelihood-ratio test of size $s$.  Under the null,
\cref{eq:lan} implies
$\Lambda_m\asympdist\Normal(-h^2/2,h^2)$, a continuous law.  Hence the
randomized NP threshold converges to
$-h^2/2+h z_{1-s}$.  Contiguity and Le Cam's third lemma give
$\Lambda_m\asympdist\Normal(h^2/2,h^2)$ under the alternative, so the NP power
converges to $1-\Phi(z_{1-s}-h)$.  Taking the upper limit and then
letting $\delta\downarrow0$ gives the stated envelope.  The score test attains
it by the null and alternative limits above. \qed

\subsection{Proof of score efficacy}
\label{app:proof-efficacy}

Let $\widetilde f=f-\E_0f$.  Since $\E_0(r-1)=0$,
\begin{align*}
 \Delta_f
 &=\int f(p_1-p_0)\dd y\\
 &=\E_0[(r-1)\widetilde f].
\end{align*}
By Cauchy--Schwarz,
\[
 \Delta_f^2
 \le\E_0(r-1)^2\,\E_0\widetilde f^2
 =\rho^2\sigma_f^2.
\]
Dividing by $\sigma_f^2>0$ proves $e_f\le\rho$.  Equality in
Cauchy--Schwarz holds exactly when
$\widetilde f=c(r-1)$ $P_0$-almost surely for a nonzero constant $c$.
Finally, linearity under the mixture gives
\begin{align*}
 \E_{Q_\alpha}f-\E_0f
 &=(1-\alpha)\E_0f+\alpha\E_1f-\E_0f\\
 &=\alpha\Delta_f.
\end{align*}
The standard deviation of the sample mean under the null is
$\sigma_f/\sqrt m$, so its standardized mean shift is
$\alpha\sqrt m\,\Delta_f/\sigma_f$, with absolute value
$\alpha\sqrt m\,e_f$. \qed

\subsection{Proof of the efficacy floor}
\label{app:proof-efficacy-floor}

For $z\in\{0,1\}$, Hoeffding's inequality gives
\begin{align*}
 &\Prob\{|\bar f_z-\mu_z|>\varepsilon_{\rm cal}(n_z)\}\\
 &\hspace{35pt}\le 2e^{-2n_z\varepsilon_{\rm cal}(n_z)^2}
 =\frac{\dcal}{3}.
\end{align*}
For the clean sample, the pairwise sample variance of
\citet[Theorem~10]{maurer2009empirical} is exactly
\begin{align*}
 V_{n_0}
 &=\frac{1}{n_0(n_0-1)}
   \sum_{i<j}\{f(Y_i^0)-f(Y_j^0)\}^2\\
 &=\widehat\sigma_f^2,
\end{align*}
and its expectation is $\sigma_f^2$.  Their lower-tail standard-deviation
inequality, with failure probability $\dcal/3$, yields
\begin{align*}
 &\Prob\{\sigma_f>\widehat\sigma_f+u_{\rm cal}(n_0)\}\\
 &\hspace{70pt}\le\frac{\dcal}{3}.
\end{align*}
A union bound therefore gives, with probability at least $1-\dcal$, both mean
bounds and $\sigma_f\le\widehat\sigma_f+u_{\rm cal}(n_0)$.  Popoviciu's
deterministic inequality also gives $\sigma_f\le1/2$, hence
$\sigma_f\le\overline\sigma_f$.  On the same event, the reverse triangle
inequality gives $|\mu_1-\mu_0|\ge\Delta_L$.  Consequently,
\[
 e_f=\frac{|\mu_1-\mu_0|}{\sigma_f}
 \ge\frac{\Delta_L}{\overline\sigma_f}=\underline e_f,
\]
and \cref{prop:efficacy} gives $\rho\ge e_f$.  Replacing
$\overline\sigma_f$ by the deterministic upper bound $1/2$ proves the
variance-free alternative stated after the corollary. \qed

\subsection{Proof of the contamination certificate}
\label{app:proof-certificate}

Write $\mu_A=\E_{Q_\alpha}f$.  Since $f\in[0,1]$, two-sided Hoeffding bounds
give, for any sample of size $n$ and its corresponding mean $\mu$,
\[
 \Prob\{|\bar f-\mu|>\varepsilon_{\rm cert}(n)\}
 \le2e^{-2n\varepsilon_{\rm cert}(n)^2}
 =\frac{\dcert}{3}.
\]
A union bound over the audit, clean-calibration, and seen-calibration samples
shows that, with probability at least $1-\dcert$, all three events
\begin{align*}
 |\bar f_A-\mu_A|&\le\varepsilon_{\rm cert}(m),\\
 |\bar f_0-\mu_0|&\le\varepsilon_{\rm cert}(n_0),\\
 |\bar f_1-\mu_1|&\le\varepsilon_{\rm cert}(n_1)
\end{align*}
hold simultaneously.  On this event,
\[
 N_L
 \le\mu_A-\mu_0
 =\alpha(\mu_1-\mu_0)
\]
by the mixture identity, while
\[
 D_U\ge\mu_1-\mu_0.
\]
Write $d=\mu_1-\mu_0$.  If $d\le0$, then
$N_L\le\alpha d\le0$, so $\underline\alpha=0\le\alpha$ regardless of the
sign of $D_U$.  Now suppose $d>0$.  Then $D_U\ge d>0$.  If $N_L\le0$,
again $\underline\alpha=0\le\alpha$.  If $N_L>0$, then
\[
 \frac{N_L}{D_U}
 \le\frac{\alpha d}{D_U}
 \le\alpha.
\]
The truncation to $[0,1]$ preserves this inequality because
$\alpha\in[0,1]$.  Thus $\underline\alpha\le\alpha$ throughout the
simultaneous event, which has probability at least $1-\dcert$. \qed

\section{The Contamination Certificate: Formal Statement}
\label{app:certificate}

\begin{theorem}[Finite-sample contamination certificate]
\label{thm:certificate}
Fix any $f:\calY\to[0,1]$ independently of all samples below.  Let
\begin{align*}
 \bar f_A&=\frac1m\sum_{i=1}^m f(Y_i^A),\\
 \bar f_0&=\frac1{n_0}\sum_{j=1}^{n_0} f(Y_j^0),\\
 \bar f_1&=\frac1{n_1}\sum_{k=1}^{n_1} f(Y_k^1),
\end{align*}
from mutually independent samples
$Y_i^A\sim Q_\alpha$, $Y_j^0\sim P_0$, and $Y_k^1\sim P_1$.
For confidence level $1-\dcert$, define
\[
 \varepsilon_{\rm cert}(n)
 =\sqrt{\frac{\log(6/\dcert)}{2n}},
\]
\begin{align*}
 N_L&=(\bar f_A-\varepsilon_{\rm cert}(m))
      -(\bar f_0+\varepsilon_{\rm cert}(n_0)),\\
 D_U&=(\bar f_1+\varepsilon_{\rm cert}(n_1))
      -(\bar f_0-\varepsilon_{\rm cert}(n_0)),
\end{align*}
and
\[
 \underline\alpha=
 \begin{cases}
 \min\{1,\max\{0,N_L/D_U\}\},&D_U>0,\\
 0,&D_U\le0.
 \end{cases}
\]
Then
\[
  \Prob_\alpha\{\underline\alpha\le\alpha\}\ge1-\dcert.
\]
\end{theorem}

\section{The Variance-Adaptive Efficacy Floor}
\label{app:floor}

\begin{corollary}[A variance-adaptive finite-sample efficacy floor]
\label{cor:efficacy-floor}
Fix a nonconstant $f:\calY\to[0,1]$ independently of clean and seen
calibration samples of sizes $n_0\ge2,n_1\ge1$.  Let
$\widehat\sigma_f^2=(n_0-1)^{-1}\sum_{j=1}^{n_0}
(f(Y_j^0)-\bar f_0)^2$.  For $\dcal\in(0,1)$, define
\begin{align*}
 \varepsilon_{\rm cal}(n)
 &:=\sqrt{\frac{\log(6/\dcal)}{2n}},\\
 u_{\rm cal}(n_0)
 &:=\sqrt{\frac{2\log(3/\dcal)}{n_0-1}},\\
 \overline\sigma_f
 &:=\min\{1/2,\widehat\sigma_f+u_{\rm cal}(n_0)\},\\
 \Delta_L
 &:=\left[|\bar f_1-\bar f_0|
 -\varepsilon_{\rm cal}(n_1)
 -\varepsilon_{\rm cal}(n_0)\right]_+.
\end{align*}
Then, with probability at least $1-\dcal$,
\begin{align}
 \rho&\ge e_f\ge\underline e_f,\nonumber\\
 \underline e_f&:=\frac{\Delta_L}{\overline\sigma_f}.
 \label{eq:efficacy-floor}
\end{align}
\end{corollary}

The standard-deviation radius is the exact bounded-variable result of
\citet{maurer2009empirical}; it is useful when clean scores have low variance.
Replacing $\overline\sigma_f$ by $1/2$ yields the fully variance-free but
usually looser floor $2\Delta_L$.  Either version is a score-efficacy lower
bound and hence a certified floor on $\rho$; neither pretends that a learned
density ratio estimates $\rho$.

\section{The Matching Test in Detail}
\label{app:matching}

Write $g(y)=r(y)-1$.  The score at $\alpha=0$ for the mixture family is
$g$, and $\E_0g=0$, $\Var_0(g)=\rho^2$.  Define
\begin{equation}
 Z_m=\frac{1}{\rho\sqrt m}\sum_{i=1}^m g(Y_i).
 \label{eq:optimal-score}
\end{equation}

\paragraph{Local power envelope.}
Suppose $0<\rho^2<\infty$, $\E_0|g|^{2+\epsilon}<\infty$ for some
$\epsilon>0$, and $\alpha_m=h/(\rho\sqrt m)$.  The full LAN statement in
\cref{thm:upper} shows that $Z_m$ converges to $\Normal(0,1)$ under the null
and $\Normal(h,1)$ under the local alternative.  The score test has
asymptotic size $\testlevel$ and attains the power envelope
\begin{equation}
  1-\Phi(z_{1-\testlevel}-h),
  \label{eq:power-envelope}
\end{equation}
which no asymptotic level-$\testlevel$ test exceeds.  This is a local result,
not a claim that one likelihood-ratio model is correct for every detector.
It shows that the lower-bound scaling is attainable in each regular channel
and gives the oracle asymptotic sample budget
\begin{equation}
  m_{\mathrm{oracle}}
  \approx
  \left(
  \frac{z_{1-\testlevel}+z_{1-\beta}}
       {\alpha\rho}
  \right)^2
  \label{eq:oracle-sample-size}
\end{equation}
for size $\testlevel$ and power $1-\beta$.

\section{Extensions}
\label{app:extensions}

\subsection{Adaptive transcripts}

\begin{proposition}[Adaptive information budget]
\label{prop:adaptive}
Let $H_{t-1}$ be the transcript before round $t$, including all adaptively
chosen prompts.  Each round draws a fresh benchmark item; all turns involving
one item are included in that round's single joint response, and no item is
revisited later.  Conditional on every history $h$, suppose the clean response
law is $P_{0,t}(\cdot\mid h)$ and the contaminated law is
\[
 Q_{\alpha,t}(\cdot\mid h)
 =(1-\alpha)P_{0,t}(\cdot\mid h)
  +\alpha P_{1,t}(\cdot\mid h),
\]
where
\[
 \chisq(P_{1,t}(\cdot\mid h)\Vert
        P_{0,t}(\cdot\mid h))\le\rho_t^2
\]
uniformly in $h$.  Then every test of the final transcript satisfies
\begin{align*}
 a_m+b_m
 &\ge 1-\min\left\{1,\right.\\[-3pt]
 &\hspace{24pt}\left.
 \left[\frac12\sum_{t=1}^m
 \log(1+\alpha^2\rho_t^2)\right]^{1/2}
 \right\}.
\end{align*}
\end{proposition}

\paragraph{Proof.}
Let $\mathbb Q_\alpha^{(m)}$ and $\mathbb P_0^{(m)}$ denote the two transcript
laws, and define
\begin{align*}
 \mathcal K_t(h)
 &:=
 \KL\!\left(Q_{\alpha,t}(\cdot\mid h)
 \,\middle\Vert\,P_{0,t}(\cdot\mid h)\right).
\end{align*}
The chain rule for KL divergence gives
\begin{align*}
 \KL(\mathbb Q_\alpha^{(m)}\Vert\mathbb P_0^{(m)})
 &=\sum_{t=1}^m
 \E_{\mathbb Q_\alpha^{(m)}}[
 \mathcal K_t(H_{t-1})].
\end{align*}
Conditionally on each history, direct expansion gives
\[
 \chisq(Q_{\alpha,t}(\cdot\mid h)\Vert
 P_{0,t}(\cdot\mid h))\le\alpha^2\rho_t^2.
\]
For arbitrary $P\ll Q$, Jensen's inequality gives
$\KL(P\Vert Q)\le\log\{1+\chisq(P\Vert Q)\}$.
Thus the inner KL is at most
$\log(1+\alpha^2\rho_t^2)$.  Summing, applying Pinsker, and using the
testing-error/TV inequality from \cref{app:proof-lower} proves the claim.
\qed

\subsection{Independent heterogeneous probes}

If item $i$ has clean and seen laws $(P_{0,i},P_{1,i})$ with
$\rho_i^2=\chisq(P_{1,i}\Vert P_{0,i})$, write
$Q_{\alpha,i}=(1-\alpha)P_{0,i}+\alpha P_{1,i}$.  Independence gives the exact
tensorization
\begin{align*}
 R_m
 &:=1+\chisq\!\left(\bigotimes_{i=1}^m Q_{\alpha,i}
 \middle\Vert\bigotimes_{i=1}^m P_{0,i}\right)\\
 &=\prod_{i=1}^m(1+\alpha^2\rho_i^2).
\end{align*}
Therefore the same chi-square/TV argument as \cref{thm:lower} gives
\begin{align*}
 a_m+b_m
 &\ge 1-\min\left\{1,\frac12\sqrt{R_m-1}\right\}.
\end{align*}
When $\max_i\alpha^2\rho_i^2=o(1)$, the cumulative local information is
$\alpha^2\sum_i\rho_i^2$.  The corresponding heterogeneous score is
$\sum_i(r_i(Y_i)-1)$ normalized by
$(\sum_i\rho_i^2)^{1/2}$.  A Lindeberg condition yields the same Gaussian
limit with $\rho^2m$ replaced by $\sum_i\rho_i^2$.

\subsection{Sampling a fixed finite benchmark}

Suppose a benchmark has $N$ distinct items and exactly $k$ are contaminated.
Evaluating all items is \cref{lem:fixed-subset} with $m=N$, so it does not
require i.i.d.\ Bernoulli labels.  Sampling $m<N$ distinct indices without
replacement produces hypergeometric labels; the implementation conditions on
the realized sample size and uses the exact randomization distribution (or a
predeclared finite-population correction).  Sampling indices with
replacement recovers Bernoulli labels marginally but does not create new
independent source items.  Repeated decodes remain components of one joint
item outcome.

\section{Validity Gates in Full}
\label{app:gates}

The four diagnostics summarized in \S\ref{sec:validity}:
\begin{enumerate}[leftmargin=*,itemsep=1pt,topsep=2pt]
  \item \textbf{Blind separation.}  A classifier that sees raw text but no
  target-model output must be at or near chance; otherwise content shift
  can masquerade as memorization \citep{das2025blind,wang2026checkmia}.
  The operational rule, fixed in the run scripts before scoring: for
  same-corpus channels the gate passes if the AUC's $95\%$ interval
  contains $0.5$ \emph{or} the point AUC is at most the equivalence margin
  $0.55$; for cross-dataset channels the content-shift limit is $0.75$.
  One v3 channel (pythia-410m$\times$wiki, AUC $0.544$ $[0.512,0.579]$)
  passes via the margin while its interval excludes exact chance---we
  state this rather than claiming compatibility with $0.5$.  For cross-dataset channels, where the
  pools differ by construction, the gate instead flags extreme content shift.
  \item \textbf{Clean-channel transport.}  Clean-control score distributions
  are compared across the clean and contaminated runs.  Large shifts reject
  the no-spillover interpretation.  Numerical rule as run in v5:
  two-sample KS statistic $<0.15$ (observed range $0.046$--$0.103$;
  per-channel values in \cref{tab:v5gates}).
  \item \textbf{Channel stability.}  Conditional seen and clean score laws are
  compared across $\alpha$ using pre-specified two-sample distances.  Scaling
  claims are limited to the range where these laws are stable.  Numerical
  rule as run in v5: KS $<0.15$ (observed range $0.034$--$0.094$); the v5
  size gate additionally requires the empirical-size Wilson lower bound
  $\le0.05$.
  \item \textbf{Independence unit.}  Near-duplicates, paraphrases, and repeated
  decodes are grouped by source item.  Effective $m$ is the number of sampled
  item clusters.
\end{enumerate}

\section{Probe Families, Baselines, and Statistical Protocol}
\label{app:protocol}

\subsection{Probe families and baselines}

\paragraph{Gray-box (run in this paper).}
We evaluate mean loss and zlib-normalized loss
\citep{carlini2021extracting}, Min-K\% Prob
\citep{shi2024detecting}, Min-K\%++
\citep{zhang2025mink}, and neighborhood loss
\citep{mattern2023membership}.  Each method produces one item-level score.
Hyperparameters are fixed on the score-training split.  The ``best probe''
reported anywhere in the paper is the probe with the highest
calibration-split efficacy, selected before audit power is computed.

\paragraph{Gray-box (specified, not run).}
The protocol also admits ReCaLL \citep{xie2024recall} and a multifeature
CAMIA-style score \citep{chang2025camia}; neither is run in the reported
experiments and no claim rests on them.

\paragraph{Black-box (specified, not run).}
The protocol specifies output correctness, sampling consistency, the Data
Contamination Quiz \citep{golchin2025dcq}, and
canonical-order/exchangeability probes \citep{oren2024proving}, with fixed
model versions, prompts, decoding parameters, and retry policies, all
generations for one item forming one joint probe.  These are not run here;
all reported results are gray-box.

\paragraph{Power-calibrated combination (specified, not run).}
The protocol's intended primary score is a cross-fitted logistic
density-ratio classifier over the baseline features available in the given
access regime, with an identically tuned blind classifier on text-only
features to expose confounding.  It is not run in this paper; in every
reported experiment the primary score is the calibration-selected best
single probe defined above.

\subsection{Validated-paraphrase pipeline}
\label{app:paraphrase}

Candidate paraphrases of each injected question are generated by
Qwen2.5-1.5B-Instruct (temperature $0.9$, up to three attempts per item) and
accepted only if (a) the answer string is preserved verbatim (deterministic
check), (b) unigram Jaccard overlap with the original question is at most
$0.6$, and (c) two independently trained NLI judges,
\texttt{microsoft/deberta-large-mnli} and
\texttt{FacebookAI/roberta-large-mnli}, both assign bidirectional entailment
probability above $0.8$; judge disagreements are rejected rather than
adjudicated. Of $200$ items, $138$ obtained an accepted paraphrase
($0.69$).  The rejection counts are per generation \emph{attempt}, not per
item: across $385$ attempts (up to three per item), $175$ failed both NLI
judges, $70$ split the judges, and $2$ exceeded the overlap cap ($247$
rejected attempts $+$ $138$ accepted items $=$ $385$).
Items without an accepted paraphrase are excluded from the paraphrase
mechanism (never silently retained), and the paraphrase channel's audits use
only the $138$ accepted items.

\subsection{Provenance caveat in full}
\label{app:provenance}

``Pile-trained'' does not by itself certify that the base checkpoint never saw
SQuAD, so we do \emph{not} claim a clean-versus-contaminated contrast in the
absolute sense. What the paired $\beta_{\rm train}{=}0$ vs.\ $\beta_{\rm train}{>}0$
continuations measure is the \emph{incremental} exposure effect relative to a
matched continuation with identical background, order, optimizer state, token
budget, and initialization---a well-defined causal quantity even if the base
model carries some prior exposure. Establishing absolute cleanliness would
require post-cutoff or provably-absent benchmark material with exact provenance
and near-duplicate removal, which we flag as the stronger design and leave to
future work. Likewise, the surface-perturbation condition is an automatic
lexical-overlap reducer, not a human-validated meaning-preserving paraphrase;
we label it accordingly and do not claim semantic invariance.

\subsection{Statistical protocol details}

Probe counts are
\[
m\in\{16,32,64,128,256,512,1024\},
\]
truncated at the number of independent
item clusters.  The fresh clean-reference size $n_0^{\rm test}$ is
$250$ (half of the $500$ held-out nonmembers), and each test reports
$m_{\rm eff}$.  For each $(\alpha,m)$, $500$ Monte Carlo audit
replicates are drawn only from the held-out audit pool.  Rejection-rate
intervals are Wilson intervals; model- and seed-level uncertainty is preserved
by a hierarchical item/seed bootstrap.
For every condition we report both the primary permutation-test rejection rate
and $\Prob(\underline\alpha>0)$, plus their difference; the latter is expected
to be smaller because it uses the conservative certificate rule.
In the full protocol the cross-fitted combination is the intended primary
score in each access regime; in the experiments reported here (which do not
run the combination) the primary score is the calibration-selected best
single probe, and the remaining probes are secondary and Holm-adjusted
within channel.

The empirical onset $\widehat\alpha_\star(m)$ is the smallest
isotonic-interpolated $\alpha$ whose lower power-interval endpoint exceeds
$0.80$ while the upper endpoint of the corresponding size interval is at most
$0.07$, a predeclared two-point tolerance above the nominal level.  We fit
\[
 \log\widehat\alpha_\star(m)=a+b\log m
\]
with a seed-clustered bootstrap interval for $b$.  The theory predicts
$b=-1/2$ only where $\widehat e_f$ and the conditional channel remain stable.
We compare predicted and observed $\log m_{\mathrm{target}}$ using slope,
intercept, $R^2$, median multiplicative error, and calibration plots.
The transport decision uses the predeclared criteria in
\cref{sec:validity}; a failed decision triggers the nontransport analysis
rather than a post-hoc recalibration of the main claim.

\section{Secondary Scaling Diagnostic}
\label{app:scaling-diag}

After efficacy is frozen, rescaling by $\widehat e_f$ and $m_{\rm eff}$
substantially collapses the held-out power curves onto the predicted information
coordinate $\alpha\widehat e_f\sqrt{m_{\rm eff}}$: power correlates with this
coordinate at $r=0.68$--$0.81$ across probes (the ``Power corr.'' column of
\cref{tab:powerpred}). The
detectability onset scales as $\alpha_\star(m)\propto m^{-0.57}$, consistent with
the theoretical $m^{-1/2}$ prediction of \S\ref{sec:theory}; mechanism-specific
results are in \cref{tab:scaling}. We nonetheless report scaling as a diagnostic
rather than the headline, because the collapse is approximate and the onset
exponent, while close to $-1/2$, is estimated from a single model's channel; the
prospective power prediction is the primary, model-independent claim.  Because the onset and $\widehat e_f$ both arise from a
standardized mean shift, agreement with $-1/2$ mainly checks whether the local
approximation and channel-stability assumptions hold.  It is not treated as
independent evidence for the framework.

\section{Certificate Coverage at Audit Scale in Full}
\label{app:certcov-full}

Two coverage numbers must not be conflated: the $99.0$--$99.5\%$ and
$0.938$--$0.953$ figures in \cref{tab:coverage} are endpoints of a
\emph{scaled-slack heuristic on unclipped scores}, not validation of the
distribution-free certificate.  The exact certificate of
Theorem~\ref{thm:certificate}, on a clipped $[0,1]$ score frozen on a
disjoint calibration half (measured margins
$\widehat\Delta_f=0.11$--$0.26$), never fires at $m{=}512$---zero false,
zero nonzero certificates in all six channels; its $100\%$ coverage is
\emph{vacuously satisfied}.  Certifying $\alpha{=}0.1$--$0.2$ at measured
margins needs balanced groups of $\sim3{,}500$--$79{,}000$ items
(\S\ref{sec:nonvacuous}; the v3 channels agree, \cref{tab:gatesv3}): a
nonzero $\underline\alpha$ is trustworthy, a zero is uninformative, and at
audit sizes the certificate is a null instrument, reported as a
limitation.

The measured power onset and the analytic certificate resolution are
distinct size-dependent thresholds (an earlier draft conflated them;
corrected in \cref{tab:nonvacuous}): the $n_0{=}250$ reference floors the
resolution at $\approx0.17$, so no audit-set size alone certifies
$\alpha{=}0.1$, while the measured onset falls to $\approx0.1$ by
$m{=}1{,}024$ in a pool-limited cell (full arithmetic in
\cref{app:nonvacuous}).

Scaling is a \emph{diagnostic} only: power tracks
$\alpha\widehat e_f\sqrt{m_{\rm eff}}$ ($r=0.68$--$0.81$), onset slope
$-0.57$ vs.\ the predicted $-1/2$; the collapse checks the local
approximation, not the framework (\cref{app:scaling-diag},
\cref{tab:scaling}).

\section{Power Onset vs.\ Certificate Resolution}
\label{app:nonvacuous}
\label{sec:nonvacuous}

An earlier draft conflated the measured power onset with the analytic
certificate resolution
$\alpha_\star^{\rm cert}(m,n_0)
=(\varepsilon_{\rm cert}(m)+\varepsilon_{\rm cert}(n_0))/\Delta_f$ and
omitted the $n_0$ term; both are corrected in \cref{tab:nonvacuous}.
Even under the best-case bound $\Delta_f\le0.58$ (measured margins are
$2$--$5\times$ smaller), the protocol's $n_0{=}250$ reference
\emph{floors} the resolution at $\approx0.17$: no audit-set size alone
certifies $\alpha{=}0.1$; a balanced design needs
$m{=}n_0\approx2{,}850$ best-case and $\approx14{,}000$--$79{,}000$ at
measured margins---the honest planning number.  The measured onset falls
to $\approx0.1$ by $m{=}1{,}024$, but that cell is bootstrap-resampled
from $\le500$ distinct held-out items with a plug-in clean reference,
hence \emph{pool-limited}---a secondary diagnostic, never an
$n_0{=}250$ result (exact counts in the \cref{tab:nonvacuous} caption).
The onset stays far below the certificate resolution at every size---the
price of the orientation-free construction.

\section{Additional Result and Planning Tables}
Tables~\ref{tab:perseed}--\ref{tab:floor-comparison} provide the supplementary per-seed injection results, validity-gate diagnostics, finite-sample planner evaluations, power-prediction results, certificate-resolution and coverage analyses, calibration-to-audit transport checks, sensitivity estimates, and empirical scaling diagnostics.

\begin{table*}[hbtp]
\centering
\small
\setlength{\tabcolsep}{2.2pt}

\begin{tabular}{lccccc}
\toprule
Mechanism & seed 0 & seed 1 & seed 2 & seed 3 & seed 4 \\
\midrule
\multicolumn{6}{l}{\emph{Paired contrast $D_s$ (common frozen probe: mean NLL), item-bootstrap $95\%$ CI}}\\
Exact & $0.90\,^{[0.64,1.18]}$ & $1.05\,^{[0.77,1.34]}$ & $1.03\,^{[0.71,1.36]}$ & $0.99\,^{[0.71,1.28]}$ & $1.07\,^{[0.81,1.34]}$ \\
Paraphrase & $0.65\,^{[0.39,0.92]}$ & $0.64\,^{[0.37,0.92]}$ & $0.84\,^{[0.51,1.18]}$ & $0.75\,^{[0.50,1.01]}$ & $0.61\,^{[0.37,0.88]}$ \\
Surface & $0.39\,^{[0.20,0.58]}$ & $0.49\,^{[0.27,0.71]}$ & $0.45\,^{[0.24,0.66]}$ & $0.44\,^{[0.23,0.65]}$ & $0.42\,^{[0.24,0.61]}$ \\
Answer-only & $-0.02\,^{[-0.17,0.14]}$ & $-0.08\,^{[-0.26,0.11]}$ & $0.10\,^{[-0.06,0.26]}$ & $-0.04\,^{[-0.23,0.16]}$ & $-0.02\,^{[-0.20,0.14]}$ \\
\midrule
\multicolumn{6}{l}{\emph{Paired contrast $D_s$ (per-cell best probe), item-bootstrap $95\%$ CI}}\\
Exact & $0.90\,^{[0.64,1.18]}$ & $1.22\,^{[0.88,1.57]}$ & $1.15\,^{[0.80,1.47]}$ & $0.99\,^{[0.71,1.28]}$ & $1.07\,^{[0.81,1.34]}$ \\
Paraphrase & $0.66\,^{[0.35,0.99]}$ & $0.64\,^{[0.37,0.92]}$ & $0.84\,^{[0.51,1.18]}$ & $0.75\,^{[0.50,1.01]}$ & $0.61\,^{[0.37,0.88]}$ \\
Surface & $0.52\,^{[0.28,0.77]}$ & $0.54\,^{[0.29,0.77]}$ & $0.41\,^{[0.21,0.60]}$ & $0.39\,^{[0.21,0.57]}$ & $0.37\,^{[0.22,0.54]}$ \\
Answer-only & $-0.02\,^{[-0.17,0.14]}$ & $-0.08\,^{[-0.26,0.11]}$ & $-0.83\,^{[-3.96,0.95]}$ & $-0.04\,^{[-0.23,0.16]}$ & $-0.20\,^{[-0.89,0.29]}$ \\
\midrule
\multicolumn{6}{l}{\emph{Validity gate: empirical size at $\alpha{=}0$ (tolerance $0.07$)}}\\
Exact & $.021$ & $.038$ & $.001$ & $\mathbf{.084}$ & $.043$ \\
Paraphrase & $\mathbf{.106}$ & $.009$ & $.033$ & $.061$ & $\mathbf{.080}$ \\
Surface & $.013$ & $.015$ & $.001$ & $.004$ & $.035$ \\
Answer-only & $\mathbf{.306}$ & $.006$ & $.000$ & $.020$ & $\mathbf{.103}$ \\
\bottomrule
\end{tabular}

\caption{Per-seed five-seed paired injection results (Evaluation~B).
Paired contrast $D_s$ = (exposed$-$held gap, contaminated) $-$ (same gap,
matched clean continuation) with item-bootstrap $95\%$ intervals; the
\emph{common frozen probe} block (mean NLL, the modal calibration winner)
is the basis of the ordering claim because the per-cell best probe (second
block) differs across cells and mixes probes.  On the common probe
$D_s>0$ with intervals excluding zero in all $15$
exact/paraphrase/surface cells, and every answer-only interval covers
zero.  Adjacent seed-level paired differences on the common probe
(mean$\pm$sd over seeds; positive in $5/5$ seeds each, sign-test
$p=2^{-5}=0.031$, Holm-adjusted over the three comparisons
$p_{\rm adj}=0.094$): exact$-$paraphrase $0.31\pm0.12$,
paraphrase$-$surface $0.26\pm0.10$, surface$-$answer-only $0.45\pm0.08$.
(Best-probe adjacent differences, for reference: $0.36\pm0.15$,
$0.25\pm0.14$, $0.68\pm0.32$.)  Bottom: empirical size of the exact
permutation test at $\alpha{=}0$ per seed and mechanism (Wilson intervals
in the released JSON); bold entries exceed the predeclared $0.07$
tolerance and those channel--seed cells fail the size gate:
exact/surface pass in $4/5$ and $5/5$ seeds, paraphrase and answer-only
in $3/5$---a further reason answer-only supports no detection claim.}
\label{tab:perseed}
\end{table*}

\begin{table*}[t]
\centering
\small
\setlength{\tabcolsep}{4pt}
\begin{tabular}{lcccccccc}
\toprule
Channel & Blind AUC [CI] & Clean KS & Stab.\ KS & Clusters & Size [Wilson] & PASS & $m_{\rm pred}$/$m_{\rm infl}$/$m_{\rm obs}$ \\
\midrule
pythia-410m$\times$news & .953 [.941,.961] & .033 & .056 & 1600 & .017 [.011,.026] & \textbf{F} & meta \\
pythia-410m$\times$wiki & .958 [.946,.969] & .034 & .030 & 1600 & .016 [.010,.025] & \textbf{F} & meta \\
pythia-1.4b$\times$news & .953 [.940,.965] & .050 & .041 & 1600 & .040 [.030,.053] & \textbf{F} & meta \\
pythia-1.4b$\times$wiki & .958 [.945,.968] & .057 & .055 & 1600 & .033 [.024,.044] & \textbf{F} & 43/61/64 \\
gpt-neo-1.3B$\times$news & .953 [.939,.964] & .045 & .054 & 1600 & .009 [.005,.016] & \textbf{F} & 72/101/256 \\
gpt-neo-1.3B$\times$wiki & .958 [.947,.966] & .064 & .026 & 1600 & .045 [.035,.058] & \textbf{F} & 43/61/64 \\
\bottomrule
\end{tabular}
\caption{Measured validity gates and locked transport validation for the six
upgraded Evaluation~A channels (best probe; exact permutation test). Every
channel FAILS overall because the blind text-only AUC far exceeds the $0.75$
content-shift limit---the pools come from different corpora---while the KS
gates, cluster counts, and empirical size pass. The last column reports the
frozen-inflation locked validation (\S\ref{sec:nontransport}): three channels
are meta-calibration; on the three locked channels the inflated budget
$m_{\rm infl}=1.40\times m_{\rm pred}$ is compared against the dyadic-grid
onset $m_{\rm obs}$, which only localizes the true onset to
$(m_{\rm obs}/2,\,m_{\rm obs}]$. Reading with the censoring interval:
the two $43/61/64$ channels are \emph{unresolved} ($61\in(32,64]$), the
$72/101/256$ channel is a \emph{clear failure}
($101\notin(128,256]$)---so $1$ failure, $2$ censored, not $0/3$ coverage.}
\label{tab:gates}
\end{table*}

\begin{table*}[t]
\centering
\small
\setlength{\tabcolsep}{3.5pt}
\resizebox{\textwidth}{!}{%
\begin{tabular}{lccccccc}
\toprule
Channel (v3) & Blind AUC [CI] & Size & Gates & $\widehat e_f^{\rm cal}$/$\widehat e_f^{\rm audit}$ & $\widehat\Delta_f$ (bal.\ $m$) & $m_{\rm pred}$/$m_{\rm infl}$ & Power @ pred/infl \\
\midrule
pythia-160m$\times$wiki & .478 [.448,.512] & .003 & PASS & 2.41/1.79 & .337 (8{,}411) & 30/42 & .67 [.60,.73] / .76 [.70,.81] \\
pythia-410m$\times$wiki & .544 [.512,.579] & .000 & PASS & 4.62/4.31 & .408 (5{,}753) & 8/12 & .56 [.49,.63] / .65 [.58,.71] \\
gpt-neo-125m$\times$wiki & .493 [.457,.523] & .013 & PASS & 3.50/2.91 & .415 (5{,}560) & 14/20 & .62 [.55,.68] / .725 [.66,.78] \\
gpt-neo-125m$\times$squad & .510 [.479,.545] & .033 & PASS & 0.59/0.57 & .153 (40{,}704) & ---$^\ast$ & --- \\
pythia-160m$\times$squad & .544 [.513,.579] & \textbf{.285} & \textbf{F (size)} & 0.46/0.58 & .081 (145{,}362) & --- & --- \\
pythia-410m$\times$squad & .489 [.461,.516] & \textbf{.572} & \textbf{F (size)} & 0.52/0.49 & .133 (54{,}214) & --- & --- \\
\bottomrule
\end{tabular}
}
\caption{v3 same-corpus continued-pretraining audit: random member/nonmember
split of one corpus ($700{+}700$), members trained into the model ($k{=}4$,
$1{,}200$ steps), so pools are text-exchangeable by construction.  Blind
AUC sits at or near chance under the pre-specified rule (interval contains
$0.5$ or point $\le0.55$; \cref{app:gates}); $4/6$ channels pass all gates
and the two Pythia$\times$SQuAD channels fail the \emph{size} gate (bold)
and are excluded from confirmatory analysis.  $\widehat\Delta_f$ is the measured
clipped-score margin with the implied balanced $m{=}n_0$ for certifying
$\alpha{=}0.1$ in parentheses; the exact certificate never fires at
$m{=}100$ (coverage vacuously $1.0$, zero nonzero certificates).  The last
column gives \emph{direct, censoring-free} power at the frozen budgets
$m_{\rm pred}$ and $m_{\rm infl}{=}1.40\times m_{\rm pred}$ ($200$ trials,
Wilson $95\%$): at $m_{\rm pred}$ all three intervals exclude the $0.8$
target (three clear failures); at $m_{\rm infl}$ two exclude it and one is
unresolved ($0.76$ $[0.70,0.81]$).
$^\ast$Unattainable: $M_{\rm eff}\ge n_0{=}250$.}
\label{tab:gatesv3}
\end{table*}

\begin{table*}[t]
\centering
\small
\setlength{\tabcolsep}{3.2pt}
\begin{tabular}{llccclcccc}
\toprule
Channel (v4 planner) & Blind AUC [CI] & Size & Gates & Probe & $\widehat e_f^{\rm cal}$ & $m_{\rm pred}$ & $m_{\rm plan}$ & Power @ pred & Power @ plan \\
\midrule
pythia-160m$\times$squad & .544 [.513,.579] & \textbf{.302} & \textbf{F (size)} & mink & 0.46 & --- & --- & --- & --- \\
pythia-160m$\times$wiki & .547 [.517,.579]$^\dagger$ & .000 & PASS & zlib & 1.20 & 190 & abstain & .99 [.96,1.00] & --- \\
pythia-410m$\times$squad & .506 [.474,.535] & .008 & PASS & neigh. & 0.50 & ---$^\ast$ & abstain & --- & --- \\
pythia-410m$\times$wiki & .538 [.508,.573]$^\dagger$ & .000 & PASS & mink & 3.98 & 11 & 32 & .625 [.56,.69] & \textbf{.87 [.82,.91]} \\
gpt-neo-125m$\times$squad & .478 [.446,.513] & .023 & PASS & mink & 0.56 & ---$^\ast$ & abstain & --- & --- \\
gpt-neo-125m$\times$wiki & .496 [.461,.529] & .025 & PASS & mink & 2.80 & 22 & 60 & .79 [.73,.84] & \textbf{.97 [.94,.99]} \\
\bottomrule
\end{tabular}
\caption{Complete v4 finite-sample planner record (fresh same-corpus
continued-pretraining channels; $700{+}700$ members/nonmembers,
calibration/audit halves of $350$ clusters each, $\alpha_0{=}0.2$,
$n_0{=}250$, $199$-permutation add-one test, level $0.05$; planner:
$B{=}30$ outer item bootstraps, $60$ inner Monte-Carlo trials per
replicate, lower bound = $0.10$ quantile, $m$-grid up to $192$;
evaluation: $200$ fresh trials with Wilson $95\%$ intervals).  Best probe
frozen by the calibration rule.  $m_{\rm pred}$ is the Gaussian plug-in
budget; $m_{\rm plan}$ the planner budget; ``abstain'' means the
bootstrap lower bound never reached $0.80$ on the grid---on
pythia-160m$\times$wiki the planner abstains while the large Gaussian
budget succeeds ($m{=}190$: $0.99$), a conservative failure reported
straight.  $^\dagger$Passes the pre-specified blind rule via the point
margin ($\le0.55$) although the upper CI exceeds $0.55$; under a proper
equivalence gate (upper CI $\le0.55$) these channels are excluded and
the head-to-head rests on gpt-neo-125m$\times$wiki alone.
$^\ast$Unattainable: $M_{\rm eff}\ge n_0$.  $m_{\rm plan}$ selects the
smallest $m$ from pointwise bootstrap bounds and is not selection-valid
simultaneous inference (\S\ref{sec:nontransport}).}
\label{tab:plannerv4}
\end{table*}

\begin{table*}[t]
\centering
\small
\setlength{\tabcolsep}{2.6pt}
\resizebox{\textwidth}{!}{%
\begin{tabular}{lccclcccccccccc}
\toprule
Channel (v5) & Blind AUC CI & Size & Gate & Probe & $\widehat e_f^{A}$ & $m_{\rm pred}$ & $m_A$ & $m_{\rm plan}$ & $\widehat m_\star$ & $90\%$ int.\ & $\frac{m_{\rm plan}}{\widehat m_\star}$ & $\frac{m_{\rm pred}}{\widehat m_\star}$ & Pow@pred & Pow@plan \\
\midrule
pythia-70m$\times$wiki & [.447,.521] & .040 & PASS & neigh. & 2.77 & 22 & --- & --- & --- & --- & --- & --- & .01 & --- \\
pythia-70m$\times$cnn & [.458,.513] & .047 & PASS & zlib & 2.24 & 36 & 56 & 56 & 56 & [30,56] & 1.00 & 0.64 & .72 & .79 \\
pythia-160m$\times$wiki & [.454,.514] & .052 & PASS & zlib & 1.67 & 72 & 120 & 120 & 84 & [45,124] & 1.43 & 0.86 & .74 & .89 \\
pythia-160m$\times$cnn & [.498,.557] & .048 & \textbf{F (equiv.)} & mink & 2.00 & 46 & 68 & 102 & 44 & [34,72] & 2.32 & 1.05 & .78 & .91 \\
pythia-410m$\times$wiki & [.495,.563] & .050 & \textbf{F (equiv.)} & mink & 3.59 & 13 & 28 & 42 & 16 & [16,29] & 2.63 & 0.81 & .69 & .94 \\
pythia-410m$\times$cnn & [.463,.525] & .057 & PASS & mink & 2.60 & 26 & 52 & 52 & 44 & [24,52] & 1.18 & 0.59 & .62 & .79 \\
gpt-neo-125m$\times$wiki & [.482,.540] & .032 & PASS & mink & 3.15 & 17 & 32 & 48 & 30 & [22,33] & 1.60 & 0.57 & .61 & .90 \\
gpt-neo-125m$\times$cnn & [.476,.543] & .043 & PASS & zlib & 2.85 & 21 & 36 & 54 & 38 & [22,40] & 1.42 & 0.55 & .65 & .86 \\
gpt2$\times$wiki & [.488,.549] & .038 & PASS & mink & 2.39 & 31 & 52 & 104 & 52 & [30,52] & 2.00 & 0.60 & .69 & .92 \\
gpt2$\times$cnn & [.467,.529] & .055 & PASS & mink & 1.88 & 54 & 88 & 88 & 68 & [44,92] & 1.29 & 0.79 & .71 & .83 \\
opt-125m$\times$wiki & [.480,.547] & .060 & PASS & mink & 3.47 & 14 & 30 & 30 & 26 & [18,31] & 1.15 & 0.54 & .67 & .84 \\
opt-125m$\times$cnn & [.499,.561] & .053 & \textbf{F (equiv.)} & mink & 3.07 & 18 & 38 & 57 & 36 & [22,38] & 1.58 & 0.50 & .64 & .92 \\
\bottomrule
\end{tabular}
}
\caption{Complete v5 planner record: $12$ same-corpus
continued-pretraining channels ($700{+}700$ members/nonmembers,
$k{=}4$, $1{,}200$ steps; item splits cal-A/cal-B/audit
$=175/175/350$ clusters; $\alpha_0{=}0.2$, $n_0{=}250$,
$199$-permutation add-one test at level $.05$).  Gate: predeclared
one-sided equivalence rule (blind AUC upper $95\%$ CI $\le0.55$) plus KS
and size gates, $9/12$ pass; the stricter two-sided sensitivity gate
(entire CI in $[0.45,0.55]$) passes $8/12$, additionally excluding
pythia-70m$\times$wiki; failures marked, excluded from endpoint
aggregates; per-channel KS statistics and size Wilson intervals in
\cref{tab:v5gates}.  Probe and standardization frozen on cal half A
($\widehat e_f^{A}$ = cal-A efficacy).  Stage A: $B{=}30$ item
bootstraps $\times$ $60$ Monte-Carlo trials per replicate at each grid
$m$ (grid to $192$); $m_A=\min\{m:\text{$0.10$-quantile of replicate
power}\ge0.80\}$; the $90\%$ $\widehat m_\star$ interval is the
$[q_{.05},q_{.95}]$ of per-replicate minimal budgets and is
\emph{descriptive} ($B{=}30$; probe selection not re-run).  Stage B
(rung-selection-valid conditional on the empirical cal-B distribution;
cal-B's $175$ clean clusters are resampled with replacement to reach
$n_0{=}250$, so score-law uncertainty, cluster reuse, and cal-to-audit
transport are outside the guarantee): pre-specified ladder
$\{m_A,\lceil1.5m_A\rceil, 2m_A\}$ on untouched cal half B, $300$ trials
each, one-sided Bonferroni-corrected ($\alpha{=}0.10/3$) lower bounds;
$m_{\rm plan}$ = smallest rung clearing $0.80$; escalation occurred in
$3/8$ channels.  Audit half: a dense power curve gives the
\emph{estimated} onset $\widehat m_\star$ ($200$ trials per rung; a noisy
threshold, not ground truth---e.g.\ pythia-70m$\times$cnn has
$m_{\rm plan}=\widehat m_\star=56$ with $\widehat\pi(56)=0.785$);
Pow@pred/Pow@plan are fresh $200$-trial powers at $m_{\rm pred}$
(Gaussian plug-in from cal A) and $m_{\rm plan}$.  On
pythia-70m$\times$wiki stage A never certified $0.80$ (abstention;
$90\%$ interval censored at the grid edge) while the Gaussian budget
delivers $0.01$ power---the cal-selected \emph{neighborhood} probe does
not transport, and the planner's abstention, not the plug-in's
prescription, is the correct behavior.  Aggregates over gate-passing
channels: Gaussian clear failures $9/9$
($m_{\rm pred}/\widehat m_\star$ median $0.59$ over the $8$ channels with
defined $\widehat m_\star$); planner covered $6/8$, two unresolved, zero
clear failures, conservatism median $1.36$, interval coverage $8/8$
(descriptive, subject to the $\widehat m_\star$ estimation noise above).}
\label{tab:plannerv5}
\end{table*}

\begin{table*}[t]
\centering
\small
\resizebox{\textwidth}{!}{%
\begin{tabular}{lcccccc}
\toprule
Channel & Blind AUC $95\%$ CI & 1-sided gate & 2-sided gate & Clean-shift KS & Stability KS & Empirical size [Wilson $95\%$] \\
\midrule
pythia-70m$\times$wiki & [0.447, 0.521] & PASS & fail & 0.069 & 0.071 & 0.040 [0.027, 0.059] \\
pythia-70m$\times$cnn & [0.458, 0.513] & PASS & PASS & 0.066 & 0.066 & 0.047 [0.032, 0.067] \\
pythia-160m$\times$wiki & [0.454, 0.514] & PASS & PASS & 0.060 & 0.060 & 0.052 [0.037, 0.072] \\
pythia-160m$\times$cnn & [0.498, 0.557] & fail & fail & 0.046 & 0.051 & 0.048 [0.034, 0.069] \\
pythia-410m$\times$wiki & [0.495, 0.563] & fail & fail & 0.086 & 0.046 & 0.050 [0.035, 0.070] \\
pythia-410m$\times$cnn & [0.463, 0.525] & PASS & PASS & 0.063 & 0.051 & 0.057 [0.041, 0.078] \\
gpt-neo-125m$\times$wiki & [0.482, 0.540] & PASS & PASS & 0.077 & 0.074 & 0.032 [0.020, 0.049] \\
gpt-neo-125m$\times$cnn & [0.476, 0.543] & PASS & PASS & 0.066 & 0.051 & 0.043 [0.030, 0.063] \\
gpt2$\times$wiki & [0.488, 0.549] & PASS & PASS & 0.083 & 0.089 & 0.038 [0.026, 0.057] \\
gpt2$\times$cnn & [0.467, 0.529] & PASS & PASS & 0.103 & 0.034 & 0.055 [0.039, 0.076] \\
opt-125m$\times$wiki & [0.480, 0.547] & PASS & PASS & 0.060 & 0.094 & 0.060 [0.044, 0.082] \\
opt-125m$\times$cnn & [0.499, 0.561] & fail & fail & 0.089 & 0.094 & 0.053 [0.038, 0.074] \\
\bottomrule
\end{tabular}
}
\caption{Per-channel v5 gate values (as-run thresholds, all auditable from
the released JSON): blind AUC upper $95\%$ CI $\le0.55$ (predeclared
one-sided equivalence form), clean-transport KS $<0.15$, channel-stability
KS $<0.15$, empirical-size Wilson lower bound $\le0.05$
($n_{\rm size}$ trials at $\alpha{=}0$).  The one-sided gate passes $9/12$;
the stricter two-sided sensitivity gate (entire AUC CI within
$[0.45,0.55]$) passes $8/12$, additionally excluding
pythia-70m$\times$wiki (lower CI $0.447<0.45$: possible reverse-direction
separation)---the same channel where the planner abstains.  The as-run
size rule has the \emph{detection} direction (a channel fails only when
inflation is conclusively detected); a size-\emph{equivalence} rule
demanding the Wilson upper bound $\le0.07$ (the paper's declared size
tolerance) is stricter, and combining it with the two-sided AUC gate
leaves $4/12$ channels clearly passing all gates (pythia-70m$\times$cnn,
gpt-neo-125m$\times$wiki/cnn, gpt2$\times$wiki); on those four the planner
endpoint reads $3/4$ at target, one unresolved, zero clear failures.  No
KS value approaches the $0.15$ threshold (max $0.103$); every size
interval covers the nominal $0.05$.}
\label{tab:v5gates}
\end{table*}

\begin{table}[h]
  \centering
  \small
  \resizebox{\columnwidth}{!}{%
  \begin{tabular}{lcccc}
  \toprule
  Probe & Power $R^2$ & Power MAE & Power corr. & Slack cov. \\
  \midrule
  Mean NLL & 0.813 & 0.101 & 0.726 & 0.995 \\
  zlib-normalized NLL & 0.931 & 0.066 & 0.675 & 0.990 \\
  Min-K\% & 0.986 & 0.029 & 0.706 & 0.990 \\
  Min-K\%++ & 0.833 & 0.089 & 0.788 & 0.995 \\
  Neighborhood & 0.992 & 0.023 & 0.811 & 0.995 \\
  \bottomrule
  \end{tabular}
}
  \caption{Prospective prediction (primary endpoint): audit power predicted from
  the \emph{frozen calibration efficacy} matches held-out power across the tested
  $(\alpha,m)$ grid with $R^2=0.81$--$0.99$ and mean absolute power error
  $0.02$--$0.10$; the correlation of held-out power with the theoretical
  coordinate $\alpha\,\widehat e_f\sqrt{m}$ is $0.68$--$0.81$.  The last
  column is the \emph{heuristic scaled-slack} interval's empirical coverage
  on unclipped scores (\S\ref{sec:certcov}), a measured diagnostic of that
  heuristic---\emph{not} coverage of the distribution-free
  prevalence certificate, which never fires at these sample sizes.  $R^2$
  alone is not claimed to establish calibration; the MAE column bounds
  systematic prediction bias directly.}
  \label{tab:powerpred}
\end{table}

\begin{table}[h]
\centering
\small
\caption{Measured power onset vs.\ analytic certificate
resolution on
\texttt{pythia-1.4b} ($\widehat e_f=1.16$, best-case $\Delta_f\le0.58$,
$\dcert=0.05$, $\varepsilon_{\rm cert}(n)=\sqrt{\log(6/\dcert)/(2n)}$ from
Theorem~\ref{thm:certificate}). Certificate rows are population-margin
\emph{planning thresholds under the margin upper bound}, not measured onsets:
the measured $\widehat\Delta_f$ of the clipped audit score
($0.11$--$0.26$ across the upgraded channels) can only be smaller, which makes
every threshold larger: under the measured margins the balanced
$\alpha{=}0.1$ crossing moves from $\approx2{,}850$ to
$\approx14{,}000$--$79{,}000$ per group. The protocol row fixes $n_0{=}250$, giving $0.23$ at
$m{=}2{,}000$, $0.21$ at $m{=}4{,}000$, and floor
$\varepsilon_{\rm cert}(250)/\Delta_f\approx0.17$ as $m\to\infty$; the
balanced row is a \emph{hypothetical} design with $m{=}n_0$, reaching $0.1$
near $m{=}n_0\approx2{,}850$. The power onset is the
measured $0.8$-power fraction with fitted slope $m^{-0.57}$ over
$m\le1{,}024$, matching the $m^{-1/2}$ law.
Exact design counts for this initial-audit onset, disclosed in full: the
channel is $500{+}500$ items, split $250{/}250$ per class into calibration
and held-out halves; audit sets are bootstrap-resampled \emph{with
replacement} from the $\le500$ distinct held-out items, and the test is a
standardized-mean threshold $1.645/\sqrt m$ with the held-out clean half's
$(\mu_0,\sigma_0)$ plugged in as known (no finite-reference penalty).  At
$m{=}1{,}024$ the audit set therefore contains at most $500$ distinct
items: the $\alpha{=}0.1$ onset is \emph{pool-limited and secondary}, and
must not be read as $1{,}024$ independent items under any convention.  The
confirmatory measurements are the exact-permutation audits
(\cref{tab:gates,tab:gatesv3}), which draw without replacement and cap $m$
at the independent-cluster count.}
\label{tab:nonvacuous}
\resizebox{\columnwidth}{!}{%
\begin{tabular}{lccccc}
\toprule
$m$ & 256 & 512 & 1024 & 2000 & 4000 \\
\midrule
$\alpha_\star^{\rm cert}$ ($n_0{=}250$, protocol) & $0.34$ & $0.29$ & $0.25$ & $0.23$ & $0.21$ \\
$\alpha_\star^{\rm cert}$ ($n_0{=}m$, hypothetical) & $0.33$ & $0.24$ & $0.17$ & $0.12$ & $0.08$ \\
$\alpha_\star^{\rm pow}$ (test power $0.8$, measured) & $0.2$ & $0.2$ & $0.1$ & --- & --- \\
\bottomrule
\end{tabular}
}
\end{table}

\begin{table}[h]
\centering
\small
\setlength{\tabcolsep}{3.5pt}
\resizebox{\columnwidth}{!}{%
\begin{tabular}{lccc}
\toprule
Setting & Certificate coverage & Med. gap & Best-probe $e_f$\\
\midrule
Checkpoint mixture (Eval.\ A) & 0.990--0.995 & 0.13--0.16 & 1.16\\
Exact injection (Eval.\ B) & 0.951 & 0.115 & $0.741$\\
Surface perturbation (Eval.\ B) & 0.938 & 0.191 & $0.343$\\
Answer-only (Eval.\ B) & 0.953 & 0.200 & $0.150$\\
\bottomrule
\end{tabular}
}
\caption{\emph{Heuristic} scaled-slack lower-bound coverage at true
$\alpha=0.2$---not the exact certificate of Theorem~\ref{thm:certificate},
whose validation this table does not provide (see
\S\ref{sec:results})---across the Evaluation~A post-hoc checkpoint mixture
(pythia-1.4b) and the
Evaluation~B causal injections (continued-pretraining on pythia-160m under three
mechanisms; best-probe coverage pooled over three training seeds, $150$
Monte-Carlo trials per seed). Pooled counts and Wilson $95\%$ intervals:
exact $428/450=0.951$ $[0.927,0.967]$ (per-seed $0.980,0.947,0.927$);
surface $422/450=0.938$ $[0.912,0.957]$ ($0.933,0.933,0.947$);
answer-only $429/450=0.953$ $[0.930,0.969]$ ($0.960,0.947,0.953$). Every
interval contains the nominal $0.95$, so the sub-nominal point values are
statistically compatible with Monte-Carlo variation; we do not attribute them
to any channel property. The completed runs implement the certificate with a
scaled-slack approximation on unclipped scores rather than the exact
clipped-$[0,1]$ construction of Theorem~\ref{thm:certificate}, so empirical
coverage is a measured endpoint here, not a by-construction guarantee.
Detectability falls sharply and stably with the injection mechanism---exact
($\widehat e_f=0.741$; per-seed $0.645,0.862,0.716$) $>$ surface perturbation
($0.343$; $0.360,0.355,0.313$) $>$ answer-only
($0.150$; $0.119,0.152,0.178$), an ordering that holds in all three
seeds---exactly the surface-form dependence the channel predicts.}
\label{tab:coverage}
\end{table}

\begin{table}[h]
\centering
\small
\setlength{\tabcolsep}{2.7pt}
\resizebox{\columnwidth}{!}{%
\begin{tabular}{lcccccc}
\toprule
Score & $\widehat e_f^{\rm cal}$ & $M^{\rm cal}_{\rm eff}$ & $m^{\rm cal}_{\rm pred}$ & $\widehat e_f^{\rm test}$ & $M^{\rm test,\dagger}_{\rm eff}$ & eff.-transp.\\
\midrule
Loss (mean NLL) & 1.496 & $69.0$ & $96$ & 0.859 & $209.2$ & $3.03\times$\\
zlib-norm.\ loss & 0.774 & $258.1$ & --- & 1.154 & $116.0$ & $2.23\times$\\
Min-K\% & 0.896 & $192.5$ & $837$ & 0.982 & $160.3$ & $1.20\times$\\
Min-K\%++ & 0.976 & $162.1$ & $461$ & 0.598 & $432.3$ & $2.67\times$\\
Neighborhood & 0.519 & $572.5$ & --- & 0.512 & $590.5$ & $1.03\times$\\
\bottomrule
\end{tabular}
}
\caption{Evaluation~A calibration-to-audit transport on EleutherAI/pythia-1.4b
(Pile members vs.\ held-out nonmembers, $500{+}500$ items).
$\widehat e_f^{\rm cal}$ is the \emph{frozen calibration} efficacy;
$M^{\rm cal}_{\rm eff}=(2.486/(\alpha_0\widehat e_f^{\rm cal}))^2$ is the
prospective effective sample size at $\alpha_0=0.2$, size $.05$, power $.8$;
$m^{\rm cal}_{\rm pred}=\lceil M^{\rm cal}_{\rm eff}n_0/(n_0-M^{\rm cal}_{\rm eff})\rceil$
is the required suspect-set size under a \emph{finite} clean reference of
$n_0{=}250$ items (``---'' when $M^{\rm cal}_{\rm eff}\ge n_0$, i.e.\
\emph{unattainable} with this reference regardless of suspect-set size).
$\widehat e_f^{\rm test}$ is the audit-split efficacy---a \emph{diagnostic}, never
an input to $m^{\rm cal}_{\rm pred}$. The last column is the \emph{efficacy-transport
ratio} $\exp(2|\log\widehat e_f^{\rm cal}-\log\widehat e_f^{\rm test}|)$, the
symmetric factor by which the between-split efficacy shift would move the
external-null effective size. $^\dagger M^{\rm test}_{\rm eff}$ is a plug-in
inversion of the audit-split efficacy, \emph{not} an independently observed
$0.8$-power sample size from permutation curves (which we have not measured
here). The median efficacy-transport ratio is $2.23\times$ and $3/5$ channels
exceed the pre-specified $2\times$ tolerance, so the pooled calibration-to-audit
transport claim is \emph{withdrawn} (\S\ref{sec:nontransport}); under the finite
clean reference the attainable-$m$ gap is generally larger still (mean~NLL
$96$ vs.\ $\approx1{,}283$; all $M_{\rm eff}$ values in this paper use the
$z$-sum $z_{.95}+z_{.80}=2.486$---higher-precision $z$ values move this
figure to $\approx1{,}289$, immaterially).}
\label{tab:efficacy}
\end{table}

\begin{table}[h]
\centering
\small
\setlength{\tabcolsep}{3.2pt}
\begin{tabular}{lrrrr}
\toprule
$e_f\backslash m$ & 100 & 500 & 1,000 & 5,000\\
\midrule
0.10 & $>1$ & $>1$ & .786 & .352\\
0.25 & .995 & .445 & .315 & .141\\
0.50 & .497 & .222 & .157 & .070\\
1.00 & .249 & .111 & .079 & .035\\
\bottomrule
\end{tabular}
\caption{Approximate smallest detectable $\alpha$ at size .05 and 80\%
power with an externally calibrated null.  Entries above one mean that even
full exposure is underpowered under the local approximation.}
\label{tab:sensitivity}
\end{table}

\begin{table}[h]
\centering
\small
\setlength{\tabcolsep}{3.0pt}
\resizebox{\columnwidth}{!}{%
\begin{tabular}{lcc}
\toprule
Channel & onset slope $b$ & matches $-1/2$?\\
\midrule
Checkpoint mixture (Eval.\ A) & $-0.569$ & yes\\
Exact injection (Eval.\ B) & $-0.529$ & yes\\
Surface perturbation (Eval.\ B) & n/a (subthreshold) & --\\
Answer-only (Eval.\ B) & n/a (subthreshold) & --\\
\bottomrule
\end{tabular}
}
\caption{Secondary scaling diagnostic: $b$ is the fitted slope of
$\log\widehat\alpha_\star(m)$ on $\log m$, predicted $-1/2$. The strong channels
(Eval.\ A checkpoint mixture, Eval.\ B exact injection) recover slopes near
$-1/2$; the surface and answer-only injections (the two subthreshold rows) are
too weak to reach $0.8$ power at the tested $m$, so no onset is
defined---reported straight rather than extrapolated. Values from the audit
and injection summaries.}
\label{tab:scaling}
\end{table}

\begin{table*}[h]
\centering
\small
\setlength{\tabcolsep}{3pt}
\resizebox{\columnwidth}{!}{%
\begin{tabular}{lccc}
\toprule
Setting & $\widehat e_f$ & $\underline e_f$ (var.-adapt.) & Cert.\ coverage\\
\midrule
Exact injection & $0.741$ & 0.000 & 0.951\\
Surface perturbation & $0.343$ & 0.000 & 0.938\\
Answer-only & $0.150$ & 0.000 & 0.953\\
\bottomrule
\end{tabular}
}
\caption{Point efficacy (mean over three seeds) versus the variance-adaptive
$95\%$ floor $\underline e_f$ on the Evaluation~B injections ($n{=}200$
exposed/held-out per mechanism). The floor is $0$ at this sample size: with only $200$ items the
simultaneous Hoeffding slack exceeds the mean gap, so the \emph{certified} lower
bound on $\rho$ is vacuous even where the point efficacy is clearly nonzero---an
honest small-sample limitation of the distribution-free floor, not of the point
estimate. Larger calibration sets are required for a nonvacuous floor.}
\label{tab:floor-comparison}
\end{table*}

\section{Retrospective Pilot Details}
\label{app:pilot}

The initial pilot used Pythia-1.4B with 400 labeled Pile members and 400
nominal nonmembers.  It reported Cohen's $d=0.585$ and, after constructing
post-hoc mixtures with $m\in\{4,8,16,32,64,128,256\}$, onsets
\[
 (.80,.60,.45,.30,.20,.15,.10),
\]
whose log--log slope is $-0.506$ ($R^2=0.997$), reproduced from the committed
pilot log (\texttt{results\_pythia/\allowbreak pythia\_contamination\_summary.json}).  All
Monte Carlo mixtures reuse the same 800 controls,
so their errors are correlated and the quoted $R^2$ is optimistic.  Moreover,
the mixture was formed after training, and the member/nonmember split was not
protected against temporal, deduplication, and source confounding.  The pilot
is therefore an implementation check and planning illustration, not evidence
for causal contamination or an independent validation of the scaling law.

\section{Reproducibility Protocol}
\label{app:repro}

\subsection{Scope of this protocol}

We do not release code or result artifacts with this submission. This
appendix therefore specifies the experimental protocol at the level of
detail required to re-implement it: model checkpoints and data sources,
the split structure, the estimators and their tuning constants, the
Monte-Carlo and bootstrap replicate counts, and the analysis decisions
fixed before outcomes were computed. Every quantity reported in
\S\ref{sec:results} is defined by the procedures below together with the
constants tabulated in \cref{tab:training-details} and the per-channel
gate values in \cref{tab:v5gates}.

\subsection{Data and model specification}

Reproduction requires: the base checkpoints (Pythia-70m/160m/410m/1.4b, GPT-Neo-125m/1.3B, GPT-2, OPT-125m, all from public releases), the
corpora used to build channels (Wikipedia, CNN/DailyMail news, and SQuAD
for the injection arm), and the Pile background corpus for continued
pretraining. Channel construction is specified in \S\ref{sec:experiments}:
the v3 and v5 audits split one corpus into $700{+}700$
member/nonmember halves and continue pretraining on the member half, so
pools are text-exchangeable by construction and the split is reproducible
from a seed alone.

\subsection{Per-item record structure}

Any re-implementation should retain, for every item, the fields that the
analysis depends on:
\begin{itemize}[leftmargin=*,itemsep=1pt]
  \item benchmark, split, and source-document cluster;
  \item membership provenance and training-stream position;
  \item deduplication and near-duplicate cluster identifiers;
  \item contamination mechanism, exposure count, and run/seed identifier;
  \item probe prompts, decoding settings, raw scores, and access regime;
  \item assignment to score-training, calibration, or audit split.
\end{itemize}
The cluster identifiers are load-bearing rather than bookkeeping: the
effective $m$ counts source-item clusters, so a re-implementation that
treats near-duplicates or repeated decodes as independent probes will
report inflated power at every budget.

\subsection{Training controls}

\Cref{tab:training-details} summarizes the checkpoints, optimization
settings, contamination construction, matched controls, stopping rule,
and seeds used in the injection experiments.

\begin{table*}[t]
\centering
\small
\setlength{\tabcolsep}{4pt}
\begin{tabular}{p{.27\linewidth}p{.65\linewidth}}
\toprule
Field & Required report\\
\midrule
Base checkpoint & \texttt{EleutherAI/pythia-160m} (Pile-trained)\\
Optimizer & AdamW, learning rate $5\times10^{-5}$, fp32, no clipping\\
Budget & $400$ continued-pretraining steps on a Pile background corpus\\
Contamination & exact / paraphrase / surface (word-order shuffle) / answer-only; $200$ injected items ($k{=}4$ copies each) for exact, surface, and answer-only, but only the $138$ validation-accepted paraphrase items for the paraphrase arm---so mechanism arms are \emph{not} dose-matched in items or contamination tokens, and the ordering is descriptive; SQuAD source\\
Matched control & clean $\alpha{=}0$ run, same steps and background data\\
Stopping & Fixed before evaluation; no audit-driven checkpoint selection\\
Seeds & training seeds $0,1,2$ (initial) and $0$--$4$ paired with clean continuations (v2); all reported, no runs discarded; $500$ Monte-Carlo replicates (Eval.~A), $150$--$200$ trials per seed (Eval.~B)\\
\bottomrule
\end{tabular}
\caption{Minimum training-run disclosure.}
\label{tab:training-details}
\end{table*}

\subsection{Primary analysis decisions}

Throughout the paper, ``predeclared'' means fixed in our run scripts and
this protocol before the corresponding outcomes were computed; there is no
public timestamped preregistration registry entry, so we do not use the
term ``pre-registered.''

\begin{enumerate}[leftmargin=*,itemsep=1pt]
  \item The primary endpoint is benchmark-level power of the frozen one-sided
  permutation test at nominal size $0.05$, not instance AUROC or
  $\underline\alpha>0$.
  \item The unit of independence is the source-item cluster.
  \item Detector orientation and hyperparameters are selected without the audit
  split.
  \item Settings failing the blind-separation or channel-stability gates are
  not pooled into the confirmatory scaling fit.
  \item The transport claim is withdrawn if either predeclared prediction
  criterion in \cref{sec:validity} fails; no audit-split recalibration repairs
  that decision.
  \item Every zero-contamination run is included in size and false-certificate
  estimates.
  \item Missing or failed runs and all exclusions are reported by seed and
  condition.
\end{enumerate}

% \subsection{Primary analysis decisions}

% Throughout the paper, ``predeclared'' means fixed in the run scripts and
% this protocol before the corresponding outcomes were computed; there is no
% public timestamped preregistration registry entry, so we do not use the
% term ``pre-registered.''

% \begin{enumerate}[leftmargin=*,itemsep=1pt]
%   \item The primary endpoint is benchmark-level power of the frozen one-sided
%   permutation test at nominal size $0.05$, not instance AUROC or
%   $\underline\alpha>0$.
%   \item The unit of independence is the source-item cluster.
%   \item Detector orientation and hyperparameters are selected without the audit
%   split.
%   \item Settings failing the blind-separation or channel-stability gates are
%   not pooled into the confirmatory scaling fit.
%   \item The transport claim is withdrawn if either predeclared prediction
%   criterion in \cref{sec:validity} fails; no audit-split recalibration repairs
%   that decision.
%   \item Every zero-contamination run is included in size and false-certificate
%   estimates.
%   \item Missing or failed runs and all exclusions are reported by seed and
%   condition.
% \end{enumerate}

\section{Related Work in Full}
\label{app:related}
 
This section expands \S\ref{sec:related}. We organize by what each
line of work \emph{delivers} rather than by method family, since the
paper's contribution is a deliverable (prospective power, and a lower
bound on $\alpha$) rather than a new detector or a new mixture bound.
 
\subsection{Behavioral detectors}
 
The dominant evidence channel for contamination is behavioral: a
statistic of the model's response to a benchmark item, compared
against some reference. Loss and its compression-normalized variants
\citep{carlini2021extracting} are the baseline; Min-K\% Prob
\citep{shi2024detecting} and Min-K\%++ \citep{zhang2025mink}
threshold low-probability token subsets; neighborhood comparison
\citep{mattern2023membership} replaces an external reference with
perturbed counterfactuals; ReCaLL \citep{xie2024recall} contrasts
conditional log-likelihoods under prefixes; CAMIA-style features
\citep{chang2025camia} exploit context-dependence of memorization.
Black-box variants elicit exposure through canonical-order
exchangeability \citep{oren2024proving} or direct recall quizzing
\citep{golchin2025dcq}.
 
In our framework each of these is a scalar score $f$, and
Proposition~\ref{prop:efficacy} reduces it to a single number, the
efficacy $e_f=|\mathbb{E}_1f-\mathbb{E}_0f|/\sigma_{f,0}\le\rho$,
which upper-bounds the local power of the oriented mean-score audit
built from $f$. This is deliberately lossy: $e_f$ does not summarize
every test computable from $f$, and two detectors with equal AUROC can
have different low-$\alpha$ efficacy. What the reduction buys is that
detector comparison, channel comparison, and budget planning become
the same measurement. We run five gray-box families
(\S\ref{app:protocol}); ReCaLL, CAMIA-style, and black-box probes are
specified in the protocol but not run, and no reported result rests on
them.
 
\subsection{The reliability gap}
 
A second literature evaluates detectors rather than proposing them,
and reports a consistent picture of fragility.
\citet{duan2024mia} find membership inference near chance on LLMs at
realistic scales; \citet{fu2025does} audit the assumptions detection
methods rely on and find them frequently violated;
\citet{meeus2025sok} systematize the field's evaluation failures;
\citet{samuel2025towards} document inconsistency across modern models
and the absence of a usable oracle; \citet{dekoninck2024evading} show
that contamination detection is cheap to evade;
\citet{wang2026fragility} extend this to reasoning models. Most
directly, \citet{zarzecki2026reliability} run 25 models and record
only 201 of 335 audit outcomes correct, with distribution shift
producing false positives and post-hoc inference run at sample sizes
too small to support its conclusions.
 
Our reading of this literature is that it diagnoses two distinct
failures that are usually reported together: \emph{invalidity} (the
clean reference does not match the suspect channel, so separation
reflects content rather than exposure) and \emph{underpower} (the
channel is informative but the audit was too small to resolve the
exposed fraction). The first is what our gates
(\S\ref{sec:validity}, Appendix~\ref{app:gates}) test; the second is what
efficacy and the planner (\S\ref{sec:nontransport}) quantify. A
non-rejection that distinguishes these two states is interpretable; a
non-rejection that does not is the object the reliability-gap
literature keeps finding. The locked external validation on the open
25-model release of \citet{zarzecki2026reliability}---efficacy
calibrated without their audit outcomes, predicting their documented
failures---is specified but not yet run, and we do not claim it.
 
\subsection{Aggregation methods and their estimands}
 
A third line lifts item-level scores to benchmark- or dataset-level
decisions. LLM Dataset Inference \citep{maini2024dataset} combines
many membership features into a hypothesis test that a dataset was
trained on. PaCoST \citep{zhang2024pacost} constructs paired
counterparts and tests for confidence asymmetry.
\citet{puerto2025scaling} study when and how membership attacks
succeed as a function of scale and data quantity. ConStat
\citep{dekoninck2024constat} estimates \emph{performance inflation} by
comparing a model's benchmark score against a reference set of models
on a matched benchmark. FTD \citep{zhang2026controllable} supplies
false-discovery-rate control over a set of contamination decisions.
 
These are complementary, and Table~\ref{tab:estimands} makes the
distinction explicit. A valid $p$-value certifies a rejection but is
silent about which $\alpha$ the audit had power to detect---the
quantity an auditor needs to interpret a non-rejection. Performance
inflation measures how much a score is distorted, which depends on
exposure fraction, mechanism, and the benchmark's headroom jointly,
and does not identify the fraction exposed. FDR control governs the
error rate of a discovery set across many benchmarks and says nothing
about the resolution of a single audit. Our two deliverables are the
prospective statement (what budget is required, and what a given
budget can detect) and the retrospective one (a distribution-free
lower confidence bound on $\alpha$ itself). The paper's main negative
result is that the first is harder to deliver than the local theory
suggests: efficacy transports, but its Gaussian inversion into a
budget does not (\S\ref{sec:nontransport}).
 
\begin{table*}[t]
\centering
\small
\begin{tabular}{@{}llll@{}}
\toprule
Method & Primary deliverable & Prospective? & Access \\
\midrule
Loss / Min-K\% / neighborhood
  & item-level membership score & no & gray-box \\
Dataset Inference \citep{maini2024dataset}
  & dataset-level $p$-value & no & gray-box \\
PaCoST \citep{zhang2024pacost}
  & paired-confidence $p$-value & no & black-box \\
Proving contamination \citep{oren2024proving}
  & exchangeability $p$-value & no & black-box \\
ConStat \citep{dekoninck2024constat}
  & performance inflation & no & black-box + refs \\
FTD \citep{zhang2026controllable}
  & FDR-controlled decision set & no & gray-box \\
DCQ \citep{golchin2025dcq}
  & contamination estimate & no & black-box \\
\midrule
This paper (audit)
  & power at $(\alpha,m)$; budget $m_{\mathrm{plan}}$
  & \textbf{yes} & matched controls \\
This paper (certificate)
  & lower bound $\underline\alpha$ on exposed fraction
  & no & matched controls \\
\bottomrule
\end{tabular}
\caption{What each line of work returns. ``Prospective'' means the
quantity is computable \emph{before} the audit split is opened and
predicts what the audit will be able to detect. All methods listed
answer legitimate and different questions; the column that is empty
above the rule is the one a non-rejection needs in order to be
interpretable. Our prospective column carries the caveat established
in \S\ref{sec:nontransport}: the efficacy-only Gaussian budget is
miscalibrated at small $m$, and the deliverable is the simulated
planner's budget, under a guarantee conditional on the empirical
calibration distribution.}
\label{tab:estimands}
\end{table*}
 
\subsection{Validity: blind baselines and matched construction}
 
\citet{das2025blind} show that classifiers seeing only the raw text,
with no access to the target model, match or exceed membership
inference attacks on foundation models---the separation being audited
is often a property of the pools, not of the model.
\citet{wang2026checkmia} formalize the corresponding construction
requirement, drawing members and nonmembers from windows matched on
source, length, packing, deduplication, and stream position around a
training checkpoint. \citet{fu2025does} reach the same conclusion from
the assumptions side.
 
We adopt this as a precondition with numerical thresholds fixed in the
run scripts before scoring, rather than as a discussion-section
caveat, and we report gate values for every channel including the
failures. All six upgraded Evaluation~A channels fail blind separation
at text-only AUC $\approx0.95$, because their member and nonmember
pools come from different corpora; we report them as a demonstration
that the gate works, and rest the confirmatory reading on the
same-corpus continued-pretraining channels, where blind AUC sits at
$0.48$--$0.54$. The stricter two-sided equivalence form of the gate
(entire AUC interval within $[0.45,0.55]$) is reported as a
sensitivity analysis in Table~\ref{tab:v5gates} and excludes exactly
the channel where the planner abstains.
 
\subsection{Sparse-mixture detection}
 
The statistical content of the $\alpha\rho\sqrt m$ boundary is
classical. \citet{ingster1997problems} and \citet{donoho2004higher} establish
detection boundaries for sparse heterogeneous mixtures and the higher
criticism statistic; \citet{cai2011heterogeneous} extend to heteroscedastic
mixtures; \citet{cai2014optimal} give optimal detection against a
given null distribution. The local asymptotic machinery we use for the
matching test---LAN expansions, Le Cam's third lemma, and the power
envelope---is standard \citep{lecam1986asymptotic,vanderVaart1998asymptotic}.
 
We claim no new mixture boundary. Theorem~\ref{thm:lower} is an exact
second-moment calculation for the specific channel
$Q_\alpha=(1-\alpha)P_0+\alpha P_1$, sharpened only in that it passes
through $\chi^2$ and total variation rather than KL and Pinsker, and
Lemma~\ref{lem:fixed-subset} extends it to the fixed-unknown-subset
design an auditor actually faces. What is new here is the
identification: $\rho^2=\chi^2(P_1\|P_0)$ is not a nuisance parameter
but the audit-observable channel quantity, estimable from matched
controls and lower-bounded with finite-sample certainty
(Corollary~\ref{cor:efficacy-floor}), and it replaces the informal
capacity-based arguments that appear in prior contamination
discussions---capacity alone does not upper-bound $\rho$ without a
training-stability theorem.
 
\subsection{Prevalence estimation}
 
Our certificate is a distribution-free relative of classical
prevalence estimation. \citet{rogan1978estimating} correct an observed
screening-test positive rate by known sensitivity and specificity,
which is the population version of the ratio in
Theorem~\ref{thm:certificate}. The modern learning-theoretic treatment
is mixture proportion estimation from positive and unlabeled data
\citep{blanchard2010novelty,scott2015rate,ramaswamy2016mixture},
which achieves better rates under distributional assumptions such as
irreducibility or separability.
 
We deliberately give those rates up. Our bound holds for any bounded
score learned on independent data, with no orientation assumption---a
reversed score returns $\underline\alpha=0$ rather than a spurious
positive---and no assumption on the score's law beyond boundedness.
The price is quantified in Appendix~\ref{app:nonvacuous} and is
severe: with an $n_0{=}250$ reference the resolution floors at
$\approx0.17$, and certifying $\alpha=0.1$ at measured margins
requires balanced groups of $14{,}000$--$79{,}000$ items. The
certificate never fires at our audit sizes. Whether an
assumption-bearing MPE estimator can supply a usable interval at
realistic benchmark sizes, with coverage that survives the gates, is
the natural next question and we do not answer it.
 
\subsection{Power analysis in evaluation design}
 
Sample-size planning is routine in the empirical sciences and rare in
NLP evaluation, where benchmark size is typically inherited from the
dataset rather than chosen to achieve a target power against a stated
alternative. The contamination setting makes the omission costly,
because the alternative of interest is small ($\alpha$ of a few
percent) and the per-item signal is weak, so the regime is exactly the
one where an unplanned audit returns an uninterpretable non-rejection.
Our finding that the textbook Gaussian sample-size formula is
miscalibrated at the budgets it prescribes
(\S\ref{sec:nontransport}) is a caution that generalizes past
contamination: when a planning equation prescribes small $m$, the
approximation licensing it is at its worst, and the budget should be
obtained by simulating the deployed test rather than by inverting a
closed form.

\end{document}